\documentclass[11pt]{article}

\usepackage[preprint]{acl}

\usepackage{times}
\usepackage{latexsym}
\usepackage[T1]{fontenc}
\usepackage[utf8]{inputenc}
\usepackage{microtype}
\usepackage{graphicx}
\usepackage{amsmath,amssymb}
\usepackage{booktabs}
\usepackage{enumitem}
\usepackage{placeins}
\usepackage{amsmath,amsfonts,bm}

\def\eqref#1{equation~\ref{#1}}

\def\1{\bm{1}}

\DeclareMathAlphabet{\mathsfit}{\encodingdefault}{\sfdefault}{m}{sl}
\SetMathAlphabet{\mathsfit}{bold}{\encodingdefault}{\sfdefault}{bx}{n}

\title{Do Halting Heads Really Save Compute\\ in Looped Language Models?}

\author{Tsung-Han Wu \and Minsuk Heo \and Kenny Tan \\
  Apple \\
  \texttt{\{th\_wu, minsuk\_heo, real\_kenny\}@apple.com}}

\begin{document}
\maketitle

\begin{abstract}
Looped language models run each token through the same layers several times, and each run is one iteration. To save compute, an extra layer, the halting head, gives after each iteration the
probability that the token stops there, so that easy tokens can stop early. Pretraining models from
scratch, up to 1.3B parameters, we find two problems. First, heads trained on a mix of all
iterations predict that tokens need under half the iterations, yet under the standard stopping rule
tokens run almost every iteration, because this training never scores the single iteration where a
token stops. Scoring every iteration instead fixes this, but in our runs it works only with a strong penalty favoring early stops. Second, savings are commonly reported as an average
number of iterations per token, but when sequences are generated in batches that run until their slowest token stops, a model without a head is almost as good at that cost. The reported average can therefore overstate the saving. We recommend reporting the iterations tokens actually run, with
the stopping threshold and batch size, and give a test, run before training a head, that bounds its gain.
\end{abstract}

\section{Introduction}
\label{sec:intro}

\begin{figure}[t]
\centering
\includegraphics[width=\columnwidth]{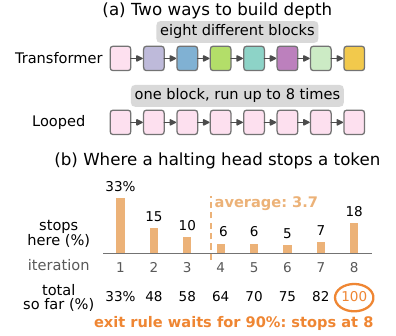}
\caption{\textbf{A head can predict $3.7$ iterations and still run all eight.} \textbf{(a)} A transformer stacks eight different blocks of layers; a looped model runs one block up to eight
times. \textbf{(b)} After each iteration, a halting head gives the probability that the token stops
there. Its average, the expected halting steps, is $1{\times}0.33 + 2{\times}0.15 + \dots +
8{\times}0.18 = 3.7$ iterations. An exit rule stops the token once the running total reaches $90\%$,
which for this illustrative token happens only at iteration 8. }
\label{fig:front}
\end{figure}

A looped language model applies one block of transformer layers to each token several
times, instead of stacking many different layers
\citep{geiping2025scaling,saunshi2025reasoning,prairie2026parcae,zeitoun2026hyperloop,fu2026simply},
so extra depth costs no extra parameters (Figure~\ref{fig:front}a). Each application of the block is
one \emph{iteration}. The number of iterations is a
setting rather than part of the trained weights: a model trained with at most eight iterations, its
\emph{cap}, can also run with fewer. This invites adaptive depth, because a model could spend two
iterations on an easy token and eight on a hard one and so save compute.

\begin{figure*}[tp]
\centering
\includegraphics[width=\textwidth]{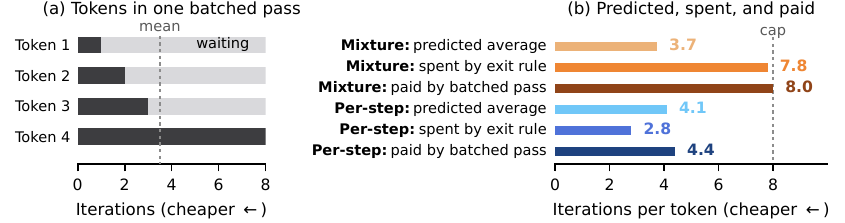}
\caption{\textbf{A batched pass costs more than the average token.} \textbf{(a)} Four tokens
processed together: the pass keeps running until its deepest token stops. \textbf{(b)} Two 370M-parameter models: the predicted average (what the head's probabilities predict), spent (what its exit rule runs) and paid (what a batched pass over $32$ sequences runs). Mixture: trained on a mixture of all iterations; per-step: every iteration scored. Thresholds as in Section~\ref{sec:bg:props}.}
\label{fig:main}
\end{figure*}

To decide when a token stops, these models add a \emph{halting head}: a small extra layer that reads
the token's state, its representation after each iteration, and outputs the probability that the token should
stop there. The two standard designs are Adaptive Computation Time (ACT)
\citep{graves2016adaptive,dehghani2018universal} and PonderNet \citep{banino2021pondernet}, and recent
looped language models use variants of both \citep{zhu2025scaling,tan2023sparse}.
Their savings are commonly reported as an average number of iterations per token. For example, Ouro
\citep{zhu2025scaling}, a released looped language model, compares its halting head with stopping
every token at one number of iterations chosen to match the average, and another model reports a
$50\%$ cut in computation, estimated from the iterations its rule skips rather than measured as
running time \citep{tan2023sparse}. This practice relies on an assumption: that the average number
of iterations per token is the compute the model spends. This paper shows that, for the halting heads we train,
this assumption fails.

Two things stand between what a halting head predicts and the compute it saves. First, the head only outputs probabilities. When the model runs, an
\emph{exit rule} decides when each token actually stops: it stops the token once the running total
of its stopping probabilities reaches a threshold, for example $90\%$. This rule stops at a
percentile of the probabilities, not at their average, the \emph{expected halting steps}.
Figure~\ref{fig:front}b shows a token whose average is $3.7$ iterations: $18\%$ of its probability
is on the last iteration, so the running total reaches $90\%$ only there, and the token runs all
eight. Second, tokens are processed in batches, and a batch keeps running until its \emph{deepest
token}, the one that runs the most iterations, has stopped, so every token in it pays for that one
(Figure~\ref{fig:main}a). Neither effect is new on its own: an average is not a percentile, and
early-exit transformers, which stop a token at one of several layers, already meet the batching
problem \citep{del2023skipdecode}. What has not been measured is how often trained heads behave like the token of Figure~\ref{fig:front}b and why, whether a different training loss fixes it, and, if it does, how much of the head's advantage survives batching.

To find out, we train halting heads into looped models of four sizes, from 140M to 1.3B
parameters, on two text corpora with three seeds for most configurations, and count the iterations
that tokens actually run, the \emph{realized steps}. Heads trained on a mixture of all their
iterations predict fewer than four of eight iterations on average, yet they run almost all eight under a $90\%$ exit rule. Stopping tokens earlier costs so much quality that, once more than about one
iteration is saved, a model of the same size with no halting head does better when run for the same
number of iterations. This model, the \emph{sampled-depth baseline}, is trained with randomly chosen
numbers of iterations.

The cause is the training loss. Like ACT, these heads are trained on a mixture of the token's states from all iterations, never on the single state at which an exit rule stops the token; we call this the \emph{mixture} loss. If the model reads that one state instead of the mixture, it becomes about $0.2$ bits per byte worse (BPB, the language-modelling loss per byte of text; lower is better). That is most of the $0.24$ BPB gap between our smallest and largest models. Scoring every iteration, as PonderNet's own \emph{per-step} loss does, is not enough on its own: it works only with a strong \emph{prior}, a training penalty that pushes stopping toward early iterations. Early-exit models already train every exit layer \citep{schuster2022confident,elhoushi2024layerskip}, but each layer's share of the loss is set without the head; the per-step loss makes each iteration's share the head's own stopping probability, and we trace the need for the prior to this weighting (Section~\ref{sec:results:recipe}).

Even then, batching removes most of the lead. Scored on whole texts at once, as in training, the per-step model beats the sampled-depth baseline run for the same number of iterations. When $32$ sequences are generated together, however, a batched pass runs more iterations than the average token, and at that cost most of the lead disappears. Batching also removes most of the saving of Ouro, a released looped model with a trained halting head. Generating one sequence at a time avoids this wait but raises a different problem: a stopped token has no keys and values at the iterations it skips, although later tokens attend to them there. The lead survives only if the token's own keys and values are reused at those iterations, as \citet{geiping2025scaling} do in their looped model. Even then, the model does not beat a \emph{specialist}, a model trained to always run about three iterations.

Our contributions are as follows:
\begin{enumerate}[leftmargin=1.2em,itemsep=1pt,topsep=2pt]
\item We show that halting heads trained with the mixture loss save almost no compute under their own exit
rule, although the heads predict that tokens need under half the iterations (Section~\ref{sec:results:earlyexit}).
\item We trace this to the loss, and show that scoring every iteration, weighted by the head's own probabilities, works only with a strong prior (Section~\ref{sec:cause}).
\item We show that batching removes most of the remaining lead, and that generating one sequence at a time keeps it only if stopped tokens' keys and values are reused (Section~\ref{sec:batching}).
\item We give a test that bounds, before any head is trained, how much a head could gain (Section~\ref{sec:feasibility}).
\end{enumerate}
Based on these results, we recommend that looped models with a halting head report three numbers: (i) the expected halting steps, (ii) the realized steps under a stated exit rule, and (iii) the iterations a batched pass runs. We also recommend stating how the keys and values of stopped tokens are filled.

\section{Background and Analysis}

\subsection{Looped models and halting heads}
\label{sec:bg:arch}

A looped model has three groups of layers (notation in Appendix~\ref{app:notation}). A \emph{prelude} of $n_p$ layers embeds the
token once. A \emph{core} of $n_r$ layers, the block of Section~\ref{sec:intro}, then runs up to $T$ times, the \emph{cap}, with the same weights. A
\emph{coda} of $n_c$ layers reads out the result. Only the core's compute grows with the number of iterations, the model's \emph{depth}.

Write $h_{i,t}$ for the state of token $i$ after $t$ iterations, that is, after the core has run $t$ times. A fixed-depth model
passes the last state, $h_{i,T}$, to the coda. A PonderNet-style halting head \citep{banino2021pondernet} instead outputs, after each iteration, the probability that token $i$ stops there if it is still running: $\lambda_{i,t} = \sigma(w^\top h_{i,t})$, where $\sigma$ is the logistic sigmoid and $w$ is the head's learned weight vector. So
$p_{i,t} = \lambda_{i,t} \prod_{u<t}(1 - \lambda_{i,u})$ is the probability that it stops exactly at
$t$, and whatever probability is left goes to the last iteration, $t = T$. Averaging the iteration number over these probabilities gives the \emph{expected halting steps}, $\mathbb{E}[N_i] = \sum_t t\, p_{i,t}$: the iteration at which the head expects the token to stop. Write $N_i$ for the iteration at which the token actually stops under the exit rule (Section~\ref{sec:bg:props}).

Given $p_i$, there are two ways to build the loss. Below, $y_i$ is the next token and $\mathcal{L}$
the cross-entropy. \textbf{The mixture loss} forms one
mixture, $\bar h_i = \sum_t p_{i,t} h_{i,t}$, and scores only the prediction $\bar y_i$ that the coda
decodes from it. \textbf{The
per-step loss}, PonderNet's own~\citep[Eq.~3]{banino2021pondernet}, scores every iteration, with the prediction $\hat y_{i,t}$ decoded
from $h_{i,t}$ alone:
\begin{align*}
\text{mixture:}\quad & \mathcal{L}(y_i, \bar y_i) + \beta\,\mathrm{KL}(p_i \,\|\, p_G),\\
\text{per-step:}\quad & \textstyle\sum_t p_{i,t}\,\mathcal{L}(y_i, \hat y_{i,t}) + \beta\,\mathrm{KL}(p_i \,\|\, p_G).
\end{align*}
In both losses, $p_G$ is a fixed prior over stopping times that favors early iterations: a geometric distribution on $t = 1, \dots, T$, $p_G(t) \propto (1 - 1/\kappa)^{t-1}$. With $\kappa{=}4$ and $T{=}8$, its mean is about three iterations. The Kullback--Leibler (KL) term pulls $p_i$ towards
$p_G$ with weight $\beta$.  We say the prior is \emph{strong} when $\beta$ is large enough to change where the head puts its
stopping probability.

\subsection{Two properties that drive the results}
\label{sec:bg:props}

\textbf{The mixture loss never scores a single iteration.} An early exit, however,
can only use the states computed so far: a single state $h_{i,t}$, or a mixture of the states up to
the exit. The loss scores neither (Appendix~\ref{app:objectives}).

\textbf{Expected halting steps is not realized compute.} Without an exit rule, the model runs all $T$
iterations: the head weights the states but does not decide how many are computed. Turning
$\mathbb{E}[N_i]$ into a saving needs a separate rule at inference, the \emph{exit rule}. The standard rule, used by ACT and by Ouro's released code, halts token
$i$ at the first $t$ where $\sum_{u \leq t} p_{i,u}$ reaches $1 - \varepsilon$. Larger $\varepsilon$
exits earlier. $\mathbb{E}[N_i]$ is an \emph{average}, but the exit rule stops
at a \emph{percentile} of the stopping probabilities: if more than $\varepsilon$ of the probability is on the last iteration, the token runs to the cap, however small its average (Figure~\ref{fig:front}b). PonderNet instead samples each token's stopping iteration from these probabilities, which spends the average but costs the mixture-loss model about $0.2$ BPB (Appendix~\ref{app:sampled}).

\paragraph{The three numbers.} We report every threshold with three numbers.
The head predicts $\mathbb{E}[N]$. The exit rule spends $\bar N$, the \emph{realized steps}: the mean, over tokens, of the
iteration $N_i$ at which the rule actually stops each token. A batched pass runs several tokens at once until its deepest token stops, so it pays that token's
iterations, written $R\,\bar N$: $R \ge 1$ is how many times more iterations the pass runs than the average token, and it grows with the number of tokens in the pass (Appendix~\ref{app:priced}). For the 370M per-step model, these are about $4.1$, $2.8$ and $4.4$ iterations (Table~\ref{tab:chain}).

\paragraph{Three ways to measure quality.} Quality is measured three ways. \emph{Teacher-forced}: the whole true text of the validation split goes through the model in one pass, as in training;
every token keeps computing until the deepest one stops, so all keys and values exist at every
iteration. This is the best case. \emph{Simulated decode}: the same one-pass computation, but a token's
keys and values are hidden at the iterations after it stops, which simulates generating one token at
a time. A \emph{real decode} checks that simulation by doing it: the model reads the true text one token at a time, as when generating, and keeps each token's keys and values in a cache, so only what
was actually computed is there. Both simulated decode and real decode can instead fill a stopped
token's missing entries with its last computed keys and values (Section~\ref{sec:batching:onetoken}). The main text reports each model at one threshold. The mixture-loss model, trained at $\beta = 0.01$, runs at $\varepsilon = 0.1$, which stops a token once $90\%$ of its probability is used; at this threshold the model keeps its quality. The per-step model, trained at $\beta = 0.1$, runs at $\varepsilon = 0.7$, which stops a token once $30\%$ of its probability is used, after about three iterations.

\begin{table}[!t]
\centering
\scriptsize
\setlength{\tabcolsep}{2pt}
\begin{tabular*}{\linewidth}{@{\extracolsep{\fill}}lccccccc@{}}
\toprule
& \multicolumn{3}{c}{\bf Iterations} & \multicolumn{4}{c}{\bf $\Delta$ BPB $\downarrow$} \\
\cmidrule(lr){2-4}\cmidrule(l){5-8}
& & & & & & \multicolumn{2}{c}{decode} \\
\bf Scale & $\mathbb{E}[N]$ & $\bar N \downarrow$ & $R\,\bar N \downarrow$ & at $\bar N$ & at $R\,\bar N$ & hidden & reused \\ \midrule
\multicolumn{8}{@{}l}{\emph{Mixture, $\varepsilon = 0.1$}} \\
140M & 3.81 & 7.86 & -- & $\mathbf{-0.0093}$ & -- & -- & -- \\
370M & 3.74 & 7.76 & 8.0 & $-0.0063$ & -- & -- & -- \\
770M & 3.63$^{\dagger}$ & 7.67 & -- & $-0.0065$ & -- & -- & -- \\
1.3B & 3.61$^{\dagger}$ & 7.61 & -- & $-0.0048$ & -- & -- & -- \\[2pt]
\multicolumn{8}{@{}l}{\emph{Per-step, $\beta{=}0.1$, $\varepsilon = 0.7$}} \\
140M & 4.33 & 3.11 & 4.71 & $\mathbf{-0.0205}$ & $-0.0004$ & $\mathbf{+0.043}$ & $\mathbf{-0.019}$ \\
370M & 4.07 & 2.83 & 4.40 & $\mathbf{-0.0219}$ & $-0.0010$ & $\mathbf{+0.035}$ & $\mathbf{-0.020}$ \\
770M & 4.00 & 2.75 & 4.37 & $\mathbf{-0.0246}$ & $\mathbf{-0.0030}$ & $\mathbf{+0.028}$ & $\mathbf{-0.022}$ \\
\bottomrule
\end{tabular*}
\caption{\textbf{What the head predicts is not what a batched pass pays.} The three \emph{Iterations} columns are the three numbers of Section~\ref{sec:bg:props}: how many iterations the head \emph{predicts} a token needs ($\mathbb{E}[N]$), how many a token \emph{spends} under the exit rule ($\bar N$), and how many a batched pass over $32$ sequences \emph{pays} ($R\,\bar N$). $\Delta$ BPB is our model's BPB minus that of the sampled-depth baseline run for the same number of iterations, so a negative value means our model is better. That baseline runs $\bar N$ iterations in the column \emph{at $\bar N$} (teacher-forced) and $R\,\bar N$ in the column \emph{at $R\,\bar N$} (a batched real decode). The mixture-loss $R\,\bar N$ is derived from $\bar N$ (Appendix~\ref{app:earlyexit}). The two \emph{decode} columns simulate generating one token at a time, and differ in what a stopped token leaves at the iterations it skipped: nothing (\emph{hidden}), or its keys and values from the iteration where it stopped (\emph{reused}). --: not measured. \textbf{Bold}: beyond noise. $^{\dagger}$One seed. }
\label{tab:chain}
\end{table}

\section{Related Work}
\begin{table*}[tp]

\centering
\footnotesize
\setlength{\tabcolsep}{3pt}
\begin{tabular*}{\linewidth}{@{\extracolsep{\fill}}llll@{}}
\toprule
\multicolumn{1}{c}{\bf Method} & \multicolumn{1}{c}{\bf Training depth}
& \multicolumn{1}{c}{\bf What the coda reads} & \multicolumn{1}{c}{\bf Loss term on depth}
\\ \midrule
Dense & no sharing, $n_p + n_r + n_c$ layers & final layer & --- \\
Sampled-depth & $T \sim$ Poisson, mean $8$ & $h_{i,T}$ & --- \\
ACT \citep{graves2016adaptive} & fixed $T{=}8$ & mixture up to its halting iteration & ponder cost, weight $\tau_p{=}10^{-4}$ \\
Mixture & fixed $T{=}8$ & mixture $\sum_t p_{i,t} h_{i,t}$ & KL to geometric $\kappa{=}4$, weight $\beta$ \\
Per-step & fixed $T{=}8$ & each $h_{i,t}$, weighted by $p_{i,t}$ & KL to geometric $\kappa{=}4$, weight $\beta$ \\
\bottomrule
\end{tabular*}
\caption{\textbf{The methods, and how each treats depth.} The sampled-depth baseline follows \citet{prairie2026parcae}. Dense has the same
number of parameters as the others but runs each layer once, so it uses fewer floating-point
operations. The ponder cost is ACT's penalty on the number of iterations a token runs (Appendix~\ref{app:actponder} varies $\tau_p$).}
\label{tab:methods}
\end{table*}

Sharing weights across layers is not new \citep{dehghani2018universal,lan2019albert}; recent work
treats the number of iterations as a variable \citep{geiping2025scaling} and studies latent reasoning \citep{saunshi2025reasoning}, scaling \citep{prairie2026parcae}, stability \citep{fu2026simply}, efficient design \citep{zeitoun2026hyperloop}, retrofitting \citep{mcleish2025teaching} and expressivity \citep{giannou2023looped}. Such sharing already makes the parameter count a misleading measure of cost \citep{dehghani2021efficiency}; an average number of iterations per token is another.

\citet{bae2025mixture} and \citet{raposo2024mixture} route tokens
to a depth or past whole blocks, with a budget fixed at training. \citet{zhu2025scaling} train looped models with a halting head that decides at each iteration whether a token stops. It is the strongest positive result for a halting head we know of, but it is compared at its realized steps, which understate the cost of a batched pass (Section~\ref{sec:batching:pass}).
\citet{tan2023sparse} train mixture-loss heads on formal-language tasks; they are
the source of the $50\%$ estimate in Section~\ref{sec:intro}. \citet{geiping2025scaling} instead stop a token without a
trained head, once successive iterations' predictions stop changing (a KL-divergence test), and
already report realized steps. The batching cost of Section~\ref{sec:batching:pass} applies to all of these. Concurrently, \citet{liu2026decodingloopedtransformersbetter} save compute without a halting head, by running every token for fewer iterations; four other concurrent papers on looped models \citep{mishra2026dynamic,popescu2026adaptive,logan2026fixedpoint,monea2026shared} are compared with our results in Appendix~\ref{app:related}. %
\citet{zhu2025scaling} train the same per-step loss under a uniform prior, then train the head alone with a separate loss on realized loss improvement. They vary the prior's shape but do not cross the loss with the prior's weight, as Section~\ref{sec:results:recipe} does.

Outside looped models, early-exit models score each exit layer, so each is a valid exit
\citep{elbayad2019depth,schuster2022confident,elhoushi2024layerskip,zhou2020bert,xin2020deebert,teerapittayanon2016branchynet,kaya2019shallow,huang2017multi},
and LayerDrop trains so that any depth can be kept \citep{fan2019reducing}.
FREE \citep{bae2023fast} fills in exited tokens' keys and values by synchronized parallel decoding, and EE-LLM \citep{chen2024eellm} recomputes them in a later pass; AdaPonderLM \citep{song2026adaponderlm} reuses halted tokens' keys and values in a looped model and reports about a tenth fewer iterations at comparable perplexity; Apparate \citep{dai2024apparate} manages early exits under batched serving. What is new here is that scoring every iteration with weights set by the head itself is not enough: it works only with a strong prior.

\section{Experimental Setup}
\label{sec:setup}
\subsection{Corpora, scales and measurement}
\label{sec:setup:data}

We pretrain from scratch on two corpora. \textbf{FineWeb-Edu (FW)}
\citep{penedo2024fineweb} is a web crawl filtered by an educational-quality classifier; the
sampled-depth baseline we build on uses it. \textbf{Common Corpus (CC)} \citep{langlais2026common} is
curated, openly licensed text. We train four
scales, from 140M to 1.3B parameters, at $76$ to $79$ tokens per parameter. In tables, $\pm$ is the standard deviation over the training seeds, and a dagger marks a cell with one seed. We report BPB on the validation split of each corpus, which training never sees: the model's cross-entropy summed over the split, in bits, divided by the split's length in bytes. Corpora, configurations, optimizer and evaluation protocol are in
Appendices~\ref{app:corpora}--\ref{app:training}, and downstream scores in Appendix~\ref{app:main}.

\paragraph{Measurement noise.} Re-evaluating the same weights on a
different GPU type moves BPB by up to $0.0045$. We call this the \emph{evaluation noise floor} and read differences between separate
evaluations against it \citep{bouthillier2021accounting}. A comparison within one evaluation session
avoids this shift, since both models move together; its difference is read against its
standard deviation across the training seeds (\emph{seed s.d.}). The per-step comparisons of
Sections~\ref{sec:cause} and~\ref{sec:batching} are made this way. In tables, \textbf{bold} marks a difference beyond twice its
noise: the seed s.d.\ for a comparison within one evaluation session, the evaluation noise floor otherwise.

\subsection{Methods compared}
\label{sec:setup:methods}

Every method in Table~\ref{tab:methods} is trained at all four scales, except two trained at three: the dense model (a standard transformer with the same prelude, core and coda layers, each run once) and the per-step model. The
two halting models, mixture-loss and per-step, use the PonderNet-style head of Section~\ref{sec:bg:arch} and
differ only in which state the coda reads; ACT stops a token once its accumulated stopping
probability reaches $1-\varepsilon$.

The sampled-depth baseline has no halting head: it is trained with a random number of iterations
(Section~\ref{sec:intro}) and run for a fixed number. We report results as \emph{margins}: a model's
BPB minus that of the sampled-depth baseline run for the same number of iterations, written $\Delta$,
so negative is better; a negative margin is a \emph{lead}. ACT and the halting models differ from that baseline in two ways:
they train at a fixed $T = 8$, not a sampled one, and they backpropagate through all eight
iterations, not four (their \emph{gradient reach}). The two halting models also add a small penalty that makes successive core states settle (Appendix~\ref{app:training}). A specialist is a model trained at one depth, chosen in each comparison to match the compared model's realized steps.

\begin{figure*}[tp]
\centering
\includegraphics[width=\textwidth]{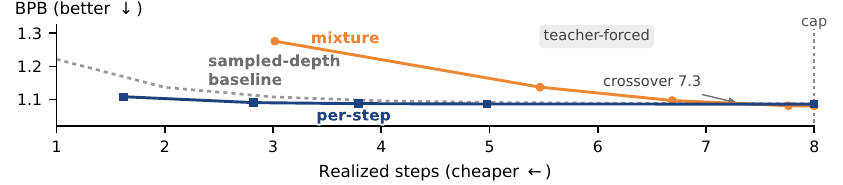}
\caption{\textbf{The cost in quality of exiting early.} BPB against realized steps,
the average number of iterations the exit rule actually runs. Each dot is one exit threshold of a 370M model. The dotted line is the sampled-depth baseline run for that many iterations; a curve below it is a lead at that cost. The mixture-loss model is above the
baseline until $7.3$ steps; the per-step model is on or below it at every threshold shown.}
\label{fig:exitcurve}
\end{figure*}

\section{Heads Trained on a Mixture Save Almost No Compute}
\label{sec:results}

In this section, we show that heads trained with the mixture loss save almost no compute under their own exit
rule, and that stopping their tokens earlier costs more quality than it saves. We first ask how much a
head could gain at all.

\subsection{How much a head could gain}
\label{sec:results:main}

A head saves compute by stopping some tokens early, so the most it can gain is set by how little quality a
model loses when it runs fewer iterations. That loss is small. At 140M, a specialist trained at three iterations is only about $0.02$ BPB worse than a specialist trained at the cap of eight. None of our models beats this eight-iteration specialist beyond noise, even when it runs all eight. Stopping tokens early is not expected to do better than running them all. So a model whose exit rule stops tokens after about three iterations can beat the three-iteration specialist by at most about $0.02$ BPB (Appendix~\ref{app:specialist}).

\subsection{Under their exit rule, tokens run almost to the cap}
\label{sec:results:earlyexit}

The exit rule stops a token once the running total of its stopping probabilities reaches $1-\varepsilon$ (Section~\ref{sec:bg:props}). A token therefore runs to the cap when more than $\varepsilon$ of its probability is on the last iteration, as the token of Figure~\ref{fig:front}b does. To see how often this happens, we apply the rule to every validation token and count the realized steps (Appendix~\ref{app:earlyexit}). At $\varepsilon = 0.1$,
the mixture-loss heads predict fewer than four of eight iterations, yet their tokens run $7.6$ to
$7.9$ (Table~\ref{tab:chain}). So the heads save almost no compute.

\paragraph{Stopping tokens earlier costs more quality than it saves.} A larger $\varepsilon$ stops tokens earlier.
Figure~\ref{fig:exitcurve} plots BPB at each threshold against its realized steps, next to the sampled-depth baseline run for the same number of iterations. At every scale the sampled-depth baseline overtakes the mixture-loss model within one iteration of the cap; Figure~\ref{fig:exitcurve} marks this point as the crossover.
So the mixture-loss model can save at most about one iteration before it falls behind.

\begin{table}[t]
\centering
\footnotesize
\setlength{\tabcolsep}{3pt}
\begin{tabular*}{\linewidth}{@{\extracolsep{\fill}}lccc@{}}
\toprule
\bf Scale & \bf \begin{tabular}[b]{@{}c@{}}Reads\\ mixture $\downarrow$\end{tabular} & \bf \begin{tabular}[b]{@{}c@{}}Reads state\\ at cap $\downarrow$\end{tabular} & \bf \begin{tabular}[b]{@{}c@{}}Readout\\ gap $\downarrow$\end{tabular} \\ \midrule
\multicolumn{4}{@{}l}{\emph{Mixture}} \\
140M & 1.2058\,\scriptsize{$\pm$.0006} & 1.4091\,\scriptsize{$\pm$.040} & $+0.203$\,\scriptsize{$\pm$.040} \\
370M & 1.0807\,\scriptsize{$\pm$.0001} & 1.2559\,\scriptsize{$\pm$.015} & $+0.175$\,\scriptsize{$\pm$.015} \\
770M & 1.0090\,\scriptsize{$\pm$.0012} & 1.1767\,\scriptsize{$\pm$.006} & $+0.168$\,\scriptsize{$\pm$.007} \\
1.3B & 0.9643\,\scriptsize{$\pm$.0017} & 1.1535\,\scriptsize{$\pm$.016} & $+0.189$\,\scriptsize{$\pm$.015} \\[2pt]
\multicolumn{4}{@{}l}{\emph{Per-step}} \\
140M & 1.2050\,\scriptsize{$\pm$.0010} & 1.2048\,\scriptsize{$\pm$.0009} & $-0.0002$\,\scriptsize{$\pm$.0001} \\
370M & 1.0924\,\scriptsize{$\pm$.0015} & 1.0914\,\scriptsize{$\pm$.0016} & $-0.0010$\,\scriptsize{$\pm$.0004} \\
770M & 1.0138\,\scriptsize{$\pm$.0003} & 1.0128\,\scriptsize{$\pm$.0003} & $-0.0010$\,\scriptsize{$<$.0001} \\
\bottomrule
\end{tabular*}
\caption{\textbf{Reading out one iteration costs the mixture-loss model about $0.2$ BPB and the per-step
model nothing.} \emph{Reads mixture} and \emph{Reads state at cap}: BPB when the coda reads the mixture or only the
state at the cap; \emph{Readout gap}: the second minus the first. All eight iterations run, so only what is
read out differs. Per-step: $\beta{=}0.01$ ($0.1$ at 770M; Appendix~\ref{app:decode}).}
\label{tab:decode}
\end{table}

\section{The Training Loss Is the Cause}
\label{sec:cause}

In this section, we show that the failure of Section~\ref{sec:results} comes from the training loss: the mixture
loss never trains the single state that an exit rule hands to the coda. We test this in three steps:
whether a single iteration can be read out, whether scoring every iteration is enough, and what a strong prior adds.

\subsection{The mixture loss never trains a single iteration}
\label{sec:results:perstep}

When the exit rule stops a token at iteration $t$, the coda reads $h_{i,t}$, or a mixture of the states up to
$t$, instead of the full mixture it was trained on. If this mismatch causes the failure, then giving the coda the state of a single iteration, instead of the mixture, should make the mixture-loss model much worse and leave the per-step model, whose loss scores every iteration, unchanged. Both hold (Table~\ref{tab:decode}). Even at the cap, with no exit, giving the coda only the last state $h_{i,T}$ costs the mixture-loss model about $0.2$ BPB, because its loss weighted that state only by the probability left for the last iteration.
One objection remains. Under early exit, the mixture-loss model reads not a single state but a mixture of the states up to the exit (Section~\ref{sec:bg:props}), so this shortened mixture, rather than the loss, could be the cause. It is not: with the same exit rule and prior weight, the per-step model beats the sampled-depth baseline at every scale (Section~\ref{sec:results:lead}), while the mixture-loss model, at two realized steps, is about $0.27$ BPB behind that baseline at 140M (Appendix~\ref{app:adaptexit}). Since the mixture-loss model fails whether the coda is given a single state (Table~\ref{tab:decode}) or the shortened mixture, the loss is the cause.

\subsection{Scoring every iteration is not enough without a strong prior}
\label{sec:results:recipe}

Every iteration of the per-step model can be read out. Yet with a weak prior ($\beta = 0.01$), it still does not stop tokens early: even at $\varepsilon = 0.95$, which stops a token once $5\%$ of its probability is used, its tokens run over $6$ of $8$ iterations (Appendix~\ref{app:pressure}).
We trace this to the
weighting: each iteration's loss is weighted by $p_{i,t}$, so training improves only the
iterations the head already favors. Only the KL term pushes depth down, and at
$\beta = 0.01$ it loses: $\mathbb{E}[N_i]$ climbs during training, well past the prior's mean of
about three (Section~\ref{sec:bg:arch}).

\paragraph{A strong prior makes the early iterations usable.} Raising $\beta$ from $0.01$ to
$0.1$ changes how quality depends on depth (Appendix~\ref{app:pressure}). At one iteration, the per-step model goes from about $0.3$ BPB \emph{worse} than the sampled-depth
baseline to about $0.06$ better, and at $\beta = 0.1$ it is never behind. We call this
\emph{depth-robustness}: the model loses less quality when it runs few iterations.

\paragraph{Neither the loss nor the prior works alone.} Table~\ref{tab:twobytwo} crosses them at one iteration, where the per-step model trails furthest. Raising $\beta$ improves the
per-step model by over $0.25$ BPB but does not help the mixture-loss model.

\begin{table}[t]
\centering
\small
\setlength{\tabcolsep}{4pt}
\begin{tabular*}{\linewidth}{@{\extracolsep{\fill}}llccc@{}}
\toprule
 & & $\beta{=}0.01$ & \multicolumn{2}{c}{change $\downarrow$ at $\beta{=}$} \\
\bf Scale & \bf Loss & BPB $\downarrow$ & $0.1$ & $1.0$ \\ \midrule
140M & Mixture & 1.813 & $+0.038$ & $+0.052$ \\
     & Per-step   & 1.711 & $\mathbf{-0.394}$ & $\mathbf{-0.450}$ \\[2pt]
370M & Mixture & 1.610 & $-0.002$ & $+0.031$ \\
     & Per-step   & 1.430 & $\mathbf{-0.273}$ & $\mathbf{-0.308}^{\dagger}$ \\
\bottomrule
\end{tabular*}
\caption{\textbf{A stronger prior helps the per-step model by over $0.25$ BPB and does not help the mixture-loss
model.} BPB on CC with every token forced to run one iteration. \textbf{Bold}: beyond noise. $^{\dagger}$One seed (Appendix~\ref{app:pressure}).}
\label{tab:twobytwo}
\end{table}

\subsection{With both, the model leads}
\label{sec:results:lead}

Unlike the mixture-loss
model, the per-step model stays ahead of the sampled-depth baseline when its exit rule stops tokens early. It spends about three iterations and beats the sampled-depth baseline run for the same number by about $0.02$ BPB, well beyond the evaluation noise floor (Table~\ref{tab:chain}).

\paragraph{The lead splits into depth-robustness and allocation.} The lead can come from two
sources.
\begin{table}[t]
\centering
\footnotesize
\setlength{\tabcolsep}{2pt}
\begin{tabular*}{\linewidth}{@{\extracolsep{\fill}}lcc@{}}
\toprule
\bf BPB $\downarrow$ & \bf Exit rule & \bf Shuffled \\ \midrule
Per-step, 140M & 1.2202 & $\mathbf{1.2285}$ \\
\bottomrule
\end{tabular*}
\caption{\textbf{Shuffling the depths the exit rule chose costs $0.008$ BPB at the same compute.}
At $\beta{=}0.1$ and $\varepsilon{=}0.7$. \emph{Exit rule}: each token runs the number of iterations the exit rule chose for it. \emph{Shuffled}: we keep the same iteration counts but hand them to the tokens in a random order, so an easy token may get the many iterations chosen for a hard one. \textbf{Bold}: beyond noise.}
\label{tab:permute}
\end{table}
The model may lose little quality at every depth, the depth-robustness of Section~\ref{sec:results:recipe}. Or its exit rule may give more iterations to the tokens that need them; we call this \emph{allocation}. To measure allocation, we shuffle the numbers of iterations the exit rule chose across tokens (Table~\ref{tab:permute}). The model then loses about two-fifths of its lead. Allocation therefore matters, but depth-robustness is most of the lead. This split, however, depends on how the sampled-depth baseline is trained. That baseline rarely trains at three or fewer iterations, so it is weak there, and part of what looks like depth-robustness may only be this weakness. We therefore retrain that baseline to draw shallower depths. At three iterations it is then nearly as good as the specialist, and the lead shrinks to about a third. Most of what remains is allocation (Appendix~\ref{app:shallow}).

\section{Batching Removes Most of the Remaining Lead}
\label{sec:batching}

Sections~\ref{sec:results} and~\ref{sec:cause} count iterations per token and score whole texts at once. In this
section, we show what changes when text is generated.

\subsection{A batched pass pays for its deepest token}
\label{sec:batching:pass}

A server generates many sequences at once, so the tokens they produce at the same time share one batched pass. The pass stops only when every token in it has stopped. It therefore costs $\max_i N_i$ iterations, not
the mean.  When $32$ sequences are
decoded together, the per-step model's tokens stop after about three iterations on average, but a pass runs $4.4$ to $4.7$ (Table~\ref{tab:chain}). A stopped token does not leave the pass. Its prediction is still the one made at the iteration where it stopped, but it keeps running the remaining iterations with the others, so it keeps writing the keys and values that later tokens attend to. Quality therefore holds; what changes is the cost. Timed at 370M with $32$ sequences decoded together, the per-step model runs about $1.6$ times as fast as eight iterations. Run at a fixed depth equal to its realized steps, the same weights would run about $2.3$ times as fast; the difference is the cost of waiting for the deepest token (Appendix~\ref{app:timing}). The lead of Section~\ref{sec:results:lead} was measured against the
sampled-depth baseline at three iterations. Compared at the iterations the pass actually runs, that
baseline is better than at three, and most of the lead disappears: what remains grows from a tie at 140M to $0.003$ BPB at 770M (Table~\ref{tab:chain}). A released model shows the same pattern. Ouro's 2.6B model \citep{zhu2025scaling} has a trained halting head; at the threshold where its BPB holds, its tokens skip about a tenth of its iterations, but a pass over $32$ tokens skips about one percent (Appendix~\ref{app:ouro}).

\paragraph{Two ways around the wait.} The first is to gather the still-running tokens into a smaller batch before each iteration, as Mixture-of-Depths routers do \citep{raposo2024mixture}; in our implementation it does not help when decoding (Appendix~\ref{app:reuse}). The second is \emph{depth-wise batching} \citep{bae2025relaxed,bae2025mixture}, which gives a stopped token's place to a token from another sequence; this works because every iteration uses the same weights. We do not test it.

\subsection{One sequence at a time, the lead needs the stopped tokens' keys and values}
\label{sec:batching:onetoken}

Decoding one sequence at a time has no deepest token to wait for. A stopped token, however, never computes its keys and values at the deeper iterations, although later tokens attend to them there. If they are left missing, the per-step model trails the sampled-depth baseline (Table~\ref{tab:chain}). A decoder can instead fill each missing entry with the
token's keys and values from the iteration where it stopped, as \citet{geiping2025scaling} do, which costs no measurable time. The model then keeps
almost all of its lead, also in a real decode (Appendix~\ref{app:reuse}).
Even then, it does not beat a specialist trained to run as many iterations as it runs on average (Appendix~\ref{app:specialist}).

\section{A Test Before Training a Head}
\label{sec:feasibility}

In this section, we give a test that bounds, before any head is trained, how much a head could gain.

Whether a head pays off depends on three quantities: $f$, the share of per-token compute inside the
core; the factor $R$ of Section~\ref{sec:bg:props}; and the \emph{depth curve}, the sampled-depth baseline's BPB at each fixed depth.
Running $t$ of $T$ iterations
costs $c(t) = (1-f) + f\,t/T$ of the full pass, so the architecture allows a speedup of $S_{\text{arch}}(t) = 1/c(t)$.

A head pays off only if two conditions hold.
\emph{Speedup}: a pass of $R\,\bar N$ iterations must cost less than one of all
$T$, $S_{\text{arch}}(R\,\bar N) > 1$.
\emph{Quality}: with its exit rule, the model must beat the sampled-depth baseline at the same $R\,\bar N$ iterations, or the saving buys a worse model.

\paragraph{The test before training.} Before any head is trained, the depth curve bounds what a head can win. A head's lead is measured against the sampled-depth baseline run for $\bar N$ iterations. Assume that stopping a token early cannot beat running it to the cap. Then the best a head can do is the quality of a model run at the cap, so its lead is at most the gap between that baseline at $\bar N$ iterations and a model at the cap. This gap has two parts. The first is how much the depth curve drops from $\bar N$ to the cap. The second is the amount by which a model trained at the cap beats that baseline at the cap, which comes from how that baseline is trained (Appendix~\ref{app:feasibility}). The first part needs no head, and the second can be measured on any model trained at the cap. So the bound is known before a head is trained. For batched decoding, the bound also needs $R$, which must then be assumed. For a head aiming at three iterations, the bound against the sampled-depth baseline is about $0.03$ BPB when one sequence is decoded at a time. When $32$ sequences are decoded together, passes run more iterations than the average token. The comparison then moves to that larger depth, where the curve has little left to drop, and the bound falls to about $0.01$.

\section{Conclusion}

For the heads we trained, the average number of iterations per token is not the compute a head saves. Heads trained with the mixture loss run tokens almost to the cap; heads scored at every iteration under a strong prior stop tokens early, but a batched pass pays for its deepest token, and at that cost most of their lead disappears. Before any head is trained, the depth curve bounds this lead (Section~\ref{sec:feasibility}).

\paragraph{What to report.} We recommend that looped models with a halting head report three things. First, the three numbers of Section~\ref{sec:bg:props}, with the exit rule's threshold and the batch size; Table~\ref{tab:chain} reports ours. Second, quality when generating: a real decode
through the key-value cache, or a simulated decode, stating how early-stopped tokens' deeper
keys and values are filled. Third, comparisons at $R\,\bar N$ iterations, not at the average, with
the sampled-depth baseline and with a specialist (Section~\ref{sec:setup:methods}). We also recommend running the test of Section~\ref{sec:feasibility} before training a head.

\section*{Limitations}
\label{sec:limitations}
\begin{itemize}[leftmargin=1.2em,itemsep=1pt,topsep=2pt]
\item \emph{The test's verdict is for a flat depth curve.} Our depth curve moves about $0.02$ BPB
over the depths an exit rule reaches; the three numbers hold whatever the curve's shape, and
only the test's verdict depends on it. Choosing how many iterations to train with, \citet{nanbeige2026} likewise find that iterations beyond two add only marginal improvement.
Training with a larger cap makes the curve steeper, but in our runs this is confounded with gradient reach and with how training depths are drawn; Appendix~\ref{app:ouro} runs the test on a curve fifteen
times steeper.

\item \emph{Evaluation is BPB on the validation split}, since the downstream aggregate cannot separate the models
(Appendix~\ref{app:main}).

\item \emph{Scale.} Our largest model has 1.3B parameters and is trained on $104$B tokens, and the per-step model is trained only up to 770M. Three claims do not depend on scale: the mixture
loss scores no single state, an exit rule stops at a percentile, not the average, and a batched pass pays for its
deepest token.

\item \emph{Seed coverage.} Table~\ref{tab:twobytwo} includes single-seed cells, and its FW replication (Appendix~\ref{app:fwinteraction}) is one seed at one scale. The baseline retrained at shallower depths (Section~\ref{sec:results:lead}) is trained at 140M only.

\item \emph{Gradient reach.} It is controlled at 140M, 370M and 770M (Appendices~\ref{app:t16}, \ref{app:feasibility}, \ref{app:fixedcontrol}).
Halting and ACT backpropagate through every iteration, the sampled-depth baseline through four, so
this difference cannot explain why an exit rule fails to stop tokens early.

\item \emph{One-token decode.} Hiding and reusing stopped tokens' keys and values are each measured in a
real decode on $64$ sequences. Why reuse works on our models is not established: the keys and values a stopped token
would compute at deeper iterations differ from its reused ones by $13$ to $26\%$, and on Ouro, at its earliest-exiting threshold, reuse did no better than keeping only the first token visible (Appendix~\ref{app:reuse}).
Gathering is timed in one implementation, without a fused kernel, and depth-wise batching (Section~\ref{sec:batching:pass}) is not tested. Training with stopped tokens' keys and values hidden, as in a one-token decode, is left to future work.

\end{itemize}

\section*{Ethical considerations}

The models here are trained from scratch for research and are not deployed; the released Ouro
checkpoints (Apache 2.0) are only evaluated. Appendices~\ref{app:corpora}--\ref{app:compute} give the configurations, data sources and evaluation protocols of every run, and the compute budget. Both corpora
are public, and we collected no personal data: CC is openly licensed text, and FW is a filtered
public web crawl, released under the Open Data Commons Attribution License (ODC-By), that may contain incidental personal information. No human subjects were involved and no annotation was
commissioned. Evaluation uses the validation splits of those corpora and the public DCLM CORE-extended suite
\citep{li2024dclm}.

The work's main cost is compute: on the order of $100{,}000$ GPU-hours (Appendix~\ref{app:compute}) for four
scales on two corpora, with three seeds for most configurations. We report it because a paper about
the true cost of a method should state its own. The test of Section~\ref{sec:feasibility} reduces
this cost: run before training a head, it can rule out a halting head and save the runs
that would otherwise establish the same thing. The reporting convention we criticize also risks provisioning deployment compute for a saving
that does not arrive.

One risk is over-generalizing these results. Section~\ref{sec:feasibility} gives the
conditions under which a halting head would pay, and the Limitations section states the scope of the
evidence.

\bibliography{refs}

\appendix
\section{Notation}
\label{app:notation}
Table~\ref{tab:notation} collects the symbols, baselines and measures of the paper, each with the
section that defines it.

\begin{table*}[!t]
\centering
\small
\setlength{\tabcolsep}{5pt}
\begin{tabular}{@{}llp{0.70\textwidth}@{}}
\toprule
\multicolumn{1}{c}{\bf Symbol} & \multicolumn{1}{c}{\bf Where} & \multicolumn{1}{c}{\bf Meaning} \\ \midrule
\multicolumn{3}{@{}l}{\emph{Model and depth}} \\
$n_p, n_r, n_c$ & \ref{sec:bg:arch} & layers in the prelude, the core and the coda; only the core repeats \\
$T$ & \ref{sec:bg:arch} & the \emph{cap}: the most iterations a token can run, set at inference \\
$T_{\text{train}}$ & Tab.~\ref{tab:configs} & the cap a model was trained at; $8$ unless stated \\
$t$ & \ref{sec:bg:arch} & an iteration, $1 \le t \le T$; one run of the core \\
$h_{i,t}$ & \ref{sec:bg:arch} & the state of token $i$ after the core has run $t$ times \\[3pt]
\multicolumn{3}{@{}l}{\emph{Halting head}} \\
$\lambda_{i,t}$ & \ref{sec:bg:arch} & the halting probability the head outputs at iteration $t$ \\
$p_{i,t}$ & \ref{sec:bg:arch} & the probability that token $i$ stops at iteration $t$; whatever probability is left goes to $t = T$ \\
$\mathbb{E}[N_i]$, $\mathbb{E}[N]$ & \ref{sec:bg:arch} & \emph{expected halting steps}, $\sum_t t\,p_{i,t}$, the average the head's probabilities predict; $\mathbb{E}[N]$ is its mean over tokens \\
$\bar h_i$ & \ref{sec:bg:arch} & the mixture $\sum_t p_{i,t} h_{i,t}$, the one state the mixture loss scores \\
mixture loss & \ref{sec:bg:arch} & scores only the mixture $\bar h_i$; a \emph{mixture-loss model} is trained with it \\
per-step loss & \ref{sec:bg:arch} & scores every $h_{i,t}$; a \emph{per-step model} is trained with it \\
attractor term & \ref{app:training} & penalty, weight $0.01$, on the squared change of successive core states; halting models only \\
$p_G$, $\kappa$ & \ref{sec:bg:arch} & geometric prior over stopping times, rate $1/\kappa$, on $\{1,\dots,T\}$; mean $3.11$ at $\kappa{=}4$, $T{=}8$ \\
$\beta$ & \ref{sec:bg:arch} & the weight on $\mathrm{KL}(p_i \,\|\, p_G)$; a \emph{strong prior} is one whose $\beta$ moves the halting mass \\[3pt]
\multicolumn{3}{@{}l}{\emph{Exit and cost}} \\
$\varepsilon$ & \ref{sec:bg:props} & the exit threshold: a token stops once its stopping probabilities sum to $1-\varepsilon$ \\
$\tau_{\text{Ouro}}$ & \ref{app:ouro} & Ouro's exit threshold: a token stops once its stopping probabilities sum to it \\
$\theta$ & \ref{app:adaptexit} & the cutoff of a per-step gate that stops a token when $\lambda_{i,t} > \theta$ \\
$N_i$, $\bar N$ & \ref{sec:bg:props} & \emph{realized steps}: the iterations token $i$ actually runs under the exit rule, and their mean \\
$R\,\bar N$, $R$ & \ref{sec:bg:props} & the iterations a batched pass pays: it runs until its deepest token stops, so $R \ge 1$ \\
$b$ & Tab.~\ref{tab:timing} & the number of sequences decoded or prefilled together in one batch \\
$B$ & \ref{app:feasibility} & the number of tokens in one batched pass: $b$ times the positions each sequence contributes \\
deepest token & \ref{sec:intro} & the token in a batched pass that runs the most iterations; the pass runs until it stops \\
depth-wise batching & \ref{sec:batching:pass} & gives a stopped token's place in the batch to another sequence's token; not tested \\
$f$ & \ref{sec:feasibility} & the share of per-token compute inside the core \\
$c(t)$, $S_{\text{arch}}$ & \ref{sec:feasibility} & the cost of $t$ of $T$ iterations relative to all $T$, $(1-f) + f\,t/T$; speedup $1/c(t)$ \\
depth curve & \ref{sec:feasibility} & the BPB of the sampled-depth baseline run at each fixed depth \\[3pt]
\multicolumn{3}{@{}l}{\emph{Training}} \\
$\mu_{\text{rec}}$ & Tab.~\ref{tab:methods} & mean of the Poisson the sampled-depth baseline draws its training depth from \\
gradient reach & \ref{sec:setup:methods} & iterations reached by gradients: all eight for ACT and halting, the last four for sampled-depth \\
$\tau_p$ & Tab.~\ref{tab:methods} & ACT's ponder weight, its penalty on computation \\
$\varepsilon_{\text{ACT}}$ & \ref{app:actponder} & ACT's exit threshold, applied to ACT's own halting mass \\
value embeddings & Tab.~\ref{tab:configs} & a token embedding added to the attention values of every other layer, as in the sampled-depth baseline \\[3pt]
\multicolumn{3}{@{}l}{\emph{Baselines and measurement}} \\
dense model & Tab.~\ref{tab:methods} & same size, not looped: the $n_p+n_r+n_c$ layers run once; trained at 140M to 770M \\
sampled-depth baseline & Tab.~\ref{tab:methods} & same size, no halting head, trained at random depths; the reference for every margin \\
specialist & \ref{sec:setup:methods} & a model trained at one fixed depth \\
shuffled depths & \ref{sec:results:lead} & the depths the exit rule chose, reassigned to random tokens; tests allocation \\
margin, $\Delta$, lead & \ref{sec:setup:methods} & a model's BPB minus the sampled-depth baseline's at equal iterations; negative is better, and a negative margin is a lead \\
BPB & \ref{sec:intro} & cross-entropy on the validation split, in bits, per byte of text; lower is better \\
validation split & \ref{sec:setup:data} & the evaluation data of each corpus, which training never sees \\
evaluation noise floor & \ref{sec:setup:data} & $0.0045$ BPB, the largest shift from re-evaluating on another device; for separate evaluations \\
seed s.d. & \ref{sec:setup:data} & s.d.\ across the three training seeds; for differences within one evaluation session \\
teacher-forced & \ref{sec:bg:props} & every position's keys and values are present at every iteration \\
simulated decode & \ref{sec:bg:props} & an exited position supplies no keys or values to later iterations, as in the cache of a decoder that runs one token per pass \\
real decode & \ref{sec:bg:props} & decoding through a real key-value cache, not emulated in one pass \\
\bottomrule
\end{tabular}
\caption{Notation, baselines and measures, with the section that defines each.}
\label{tab:notation}
\end{table*}

\section{Setup and reproduction}

\subsection{Corpora}
\label{app:corpora}

\paragraph{FineWeb-Edu (FW).} We train on a shuffled copy of the \texttt{sample-100BT} split of
FW \citep{penedo2024fineweb}, released with the sampled-depth baseline \citep{prairie2026parcae}, and evaluate on the
validation split released with it.

\paragraph{Common Corpus (CC).} We use the \emph{Open Culture}, \emph{Open Science}
and \emph{Semantic data} collections of \texttt{PleIAs/common\_corpus} \citep{langlais2026common}: curated,
openly licensed text drawn from public-domain books, scientific writing and
structured knowledge sources. We drop the other three: Open Web, which may
contain Common Crawl data, and Open Government and Open Code, to keep the corpus to books, science
and structured knowledge. Training uses a fixed snapshot
re-chunked into $864$ shards, also released with the sampled-depth baseline; the validation split ($2{,}047{,}000$ tokens) is kept out of it and
is the same for every run.

Both corpora use the sampled-depth baseline's tokenizer (vocabulary $32{,}768$) and $2048$-token
sequences. Both are public Hugging Face datasets.

\subsection{Model configurations}
\label{app:configs}
The four looped models differ in loss, cap and gradient reach (Table~\ref{tab:methods}; gradient reach, Section~\ref{sec:setup:methods}); architecture, data and optimizer are shared.
All use the prelude--core--coda stack of Section~\ref{sec:bg:arch}, with $n_p = n_r = n_c$ and RMSNorm
throughout; Table~\ref{tab:configs} gives widths, layer counts and token budgets.

\subsection{Training and evaluation protocol}
\label{app:training}

\paragraph{Why over-train.} $76$--$79$ tokens per parameter is well past the compute-optimal point, the token count that gives the best model for a fixed training budget (about $20$; \citealp{hoffmann2022training,besiroglu2024chinchilla}). It is the regime that inference-aware scaling favors, where a model that will be run many times is better trained longer at a smaller size \citep{sardana2023beyond}.

\paragraph{The baseline.} The sampled-depth baseline uses its released recipe \citep{prairie2026parcae} unchanged.

\paragraph{Seeds.} Every configuration is trained with seeds $7$, $42$ and $123$ unless a table marks a cell with one seed (a dagger).

\paragraph{The attractor term.} Both halting losses add a penalty on the squared change between
successive core states, at weight $0.01$. The weight was chosen on the mixture-loss model from
$\{0, 0.01, 1.0\}$. At $1.0$ two 140M CC seeds diverged (BPB $1.99$, against $1.2052$ at $0.01$). Without the
term, CC BPB at $T{=}8$ moves by at most $0.0016$ at 370M, 770M and 1.3B (three seeds each), inside
the evaluation noise floor, and at 140M two of three seeds train worse (mean $1.2295$ against $1.2052$).

\paragraph{The ACT ponder cost.} $\tau_p = 10^{-4}$ was chosen at 140M on CC by BPB at the training
depth, from $\{10^{-4},\, 3\times10^{-3},\, 10^{-2}\}$, three seeds each, and is used at every scale.
Appendix~\ref{app:actponder} extends the sweep to weights where ACT's depth varies across tokens.

\paragraph{Optimization.} All runs use MuonAdamW
\citep{jordan2024muon,liu2025muon}, which applies Muon to hidden matrices and
AdamW \citep{kingma2014adam,loshchilov2017decoupled} to the remaining parameters.
The AdamW group uses peak learning rate $8\times10^{-3}$ at 140M and 370M and
$6\times10^{-3}$ at 770M and 1.3B, $\beta = (0.8, 0.95)$, $\epsilon = 10^{-10}$ and
no weight decay; the Muon group uses learning rate $8\times10^{-3}$ at every scale,
momentum $0.95$, weight decay $0.2$ and $5$ Newton--Schulz iterations. The schedule
is trapezoidal \citep{hu2024minicpm,hagele2024scaling} with \emph{no} warmup: the
learning rate is held at its peak for the first half of training and decayed
linearly to zero over the second half. Gradients are clipped at $1.0$ and
non-finite gradients are skipped. The world batch is $256$ sequences of $2048$
tokens ($524{,}288$ tokens per optimizer step); the micro-batch varies from $8$ to
$32$ with the device count, with accumulation holding the world batch fixed. Token
budgets are in Table~\ref{tab:configs}.

\begin{table}[!t]

\begin{center}
\small
\setlength{\tabcolsep}{2pt}
\begin{tabular*}{\linewidth}{@{\extracolsep{\fill}}lccccc@{}}
\toprule
\multicolumn{1}{c}{\bf Scale} & \multicolumn{1}{c}{\bf Params}
& \multicolumn{1}{c}{\bf $d_{\text{model}}$}
& \multicolumn{1}{c}{\bf $d_{\text{ffn}}$} & \multicolumn{1}{c}{\bf $n_p/n_r/n_c$}
& \multicolumn{1}{c}{\bf Tokens} \\ \midrule
140M  & 144.3M & 768  & 3072 & 2/2/2 & 11.2B \\
370M  & 388.0M & 1024 & 4096 & 4/4/4 & 29.6B \\
770M  & 776.7M & 1280 & 5120 & 6/6/6 & 61.6B \\
1.3B  & 1.339B & 1536 & 6144 & 8/8/8 & 104B \\
\bottomrule
\end{tabular*}
\end{center}
\caption{Model configurations. Looped models train at $T_{\text{train}}=8$ unless stated
(Table~\ref{tab:methods}); the dense models (140M--770M) run the same $n_p+n_r+n_c$ layers once,
without looping. Parameter counts are exact and include the sampled-depth baseline's value embeddings (Appendix~\ref{app:notation}), 45--52\% of parameters; output
embeddings are tied. Scale names follow that baseline's. $d_{\text{model}}$: the width of the hidden state; $d_{\text{ffn}}$: the width of the MLP's hidden layer.}
\label{tab:configs}
\end{table}

\paragraph{Evaluation.} We report BPB on the validation split and, as a downstream aggregate, the CORE-extended score of DCLM (DataComp for Language Models; \citealp{li2024dclm}): $46$ tasks, each rescaled so that random guessing scores zero, scored on the
published eval bundle \citep{karpathy2025nanochat} so the task set is fixed. These values are not
directly comparable to published CORE numbers centered on the revised DCLM-v2 baselines
\citep{wang2026smelt}.

\raggedbottom
\subsection{Compute budget}
\label{app:compute}

The project used on the order of $100{,}000$ GPU-hours by the scheduler's accounting, on preemptible
H100, H200 and B200 instances (mostly H200), including evaluation and exploratory runs. For scale, one 1.3B sampled-depth baseline run takes about five days on one
$8{\times}$B200 node, roughly $960$ GPU-hours; halting runs take longer, since they backpropagate
through every iteration. In total we pretrained more than $100$ models across the four scales.

\section{The cost model and the speedup ceiling}

\subsection{Wall-clock timing}
\label{app:timing}

This appendix measures how much faster a pass becomes when the core runs fewer iterations, and how much of that speedup the exit rule delivers. A pass at one iteration is about three times faster than at eight (Table~\ref{tab:timing}); under the exit rule the speedup is smaller than at a fixed depth equal to the realized steps, mostly because a batched pass waits for its deepest token (Table~\ref{tab:synchrony}).

Unless a block header says otherwise, all numbers here are measured on one exclusive NVIDIA H200, with
the body in bfloat16 (a 16-bit number format) and the halting head in 32-bit floating point, as trained, and are medians over nine repeats.
We quote ratios rather than absolute times. Prefill is one pass over $2048$ positions per sequence;
decode generates $64$ tokens through a key-value cache after a $512$-token prompt. Each row states $b$, the
number of sequences per pass.

\flushbottom
\paragraph{What an exited token contributes to later tokens.} A pass runs the core on every
position until no position is still active: exit fixes which state a position is read from, not
whether it keeps computing. In the teacher-forced evaluations, later positions therefore attend to
keys and values at every depth the pass reaches, including the deeper iterations of positions that
have already exited. Decoding one token per pass differs: the loop stops when that token exits, so
the cache never receives its deeper keys and values, and no states are copied in their place as in
\citet{schuster2022confident}. Later tokens then attend to fewer keys at deep iterations, but
attention is only $L/(6d+L)$ of core cost ($L$ the context length, $d = d_{\text{model}}$ the width: attention takes about $2Ld$ multiply-adds per token per layer, the projections and the MLP about $12d^2$; $5.9$--$11.1\%$ at
our decode context), so this saves little compute. Appendix~\ref{app:decodematched} measures quality
in this setting, where the missing keys and values are hidden. A decoder can instead fill them: it reuses the token's keys and values from the iteration where it stopped (Section~\ref{sec:batching:onetoken}, Appendix~\ref{app:reuse}).

\paragraph{The speedup ceiling.} Speedup grows \emph{sublinearly} as the iteration count falls
(Table~\ref{tab:timing}): $T=1$ gives $2.8$--$3.3\times$, not $8\times$. The prelude, coda and
unembedding are a fixed cost, so the core is $73$--$79\%$ of the full pass
(Appendix~\ref{app:feasibility} fits this share row by row). At $T{=}8$, about $3\times$ therefore
bounds any scheme that varies only the core's iteration count.

\begin{table*}[!t]

\begin{center}
\footnotesize
\setlength{\tabcolsep}{4pt}
\begin{tabular*}{\textwidth}{@{\extracolsep{\fill}}llcccccccc@{}}
\toprule
\multicolumn{1}{c}{\bf Scale} & \multicolumn{1}{c}{\bf Workload}
& \multicolumn{1}{c}{$T{=}1$} & \multicolumn{1}{c}{$T{=}2$}
& \multicolumn{1}{c}{$T{=}3$} & \multicolumn{1}{c}{$T{=}4$}
& \multicolumn{1}{c}{$T{=}5$} & \multicolumn{1}{c}{$T{=}6$}
& \multicolumn{1}{c}{\bf Fitted $f$} & \multicolumn{1}{c}{\bf Largest error}
\\ \midrule
140M  & prefill, $b{=}32$ & 2.76 & 2.21 & 1.84 & 1.57 & 1.38 & 1.22 & $0.729$ & $0.005$ \\
370M  & prefill, $b{=}32$ & 3.05 & 2.36 & 1.92 & 1.62 & 1.40 & 1.24 & $0.768$ & $0.005$ \\
770M  & prefill, $b{=}32$ & 3.17 & 2.42 & 1.96 & 1.64 & 1.42 & 1.24 & $0.782$ & $0.005$ \\
1.3B  & prefill, $b{=}32$ & 3.22 & 2.45 & 1.97 & 1.65 & 1.42 & 1.25 & $0.788$ & $0.005$ \\[3pt]
140M  & decode, $b{=}1$   & 3.16 & 2.42 & 1.95 & 1.65 & 1.41 & 1.24 & $0.781$ & $0.009$ \\
370M  & decode, $b{=}1$   & 3.24 & 2.43 & 1.92 & 1.61 & 1.41 & 1.24 & $0.790$ & $0.056$ \\
770M  & decode, $b{=}1$   & 3.26 & 2.47 & 1.95 & 1.67 & 1.44 & 1.24 & $0.792$ & $0.031$ \\
1.3B  & decode, $b{=}1$   & 3.24 & 2.47 & 1.99 & 1.66 & 1.38 & 1.22 & $0.790$ & $0.041$ \\
\bottomrule
\end{tabular*}
\end{center}
\caption{\textbf{The architectural speedup ceiling.} Forced-depth speedup against $T=8$, median of
nine repeats. Run-to-run variation is at most $0.070\times$ (median $0.010\times$). Across the four scales at one
$T$, speedups differ by under $0.2\times$ in decode, and also in prefill for $T \ge 3$. \emph{Fitted $f$}: the cost model's one parameter (Section~\ref{sec:feasibility}), fitted to the row's $T{=}1$ speedup; \emph{Largest error}: its largest error in predicting the row's $T = 2$ to $6$ speedups (Appendix~\ref{app:feasibility}).}
\label{tab:timing}
\end{table*}

\begin{figure}[!t]
\begin{center}
\includegraphics[width=\columnwidth]{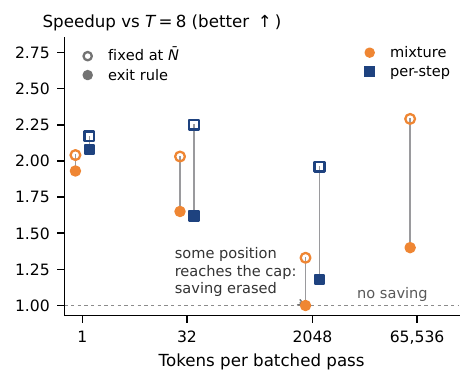}
\end{center}
\caption{\textbf{A batched pass costs the maximum, not the mean.} Measured speedup over $T{=}8$ at
370M: the exit rule (filled) beside the same weights at a fixed depth equal to their realized steps
(open, interpolated). Thresholds differ between points (Table~\ref{tab:synchrony}): compare filled
with open at one $x$ only. Per-step prefill at $b{=}32$ was not timed at 370M. The $x$-axis is
tokens per batched pass (sequences $\times$ positions). This count, not the number of sequences, decides how deep the deepest token goes.}
\label{fig:synchrony}
\end{figure}

\paragraph{What Figure~\ref{fig:synchrony} measures.} At 370M, decoding one sequence, one token per pass,
is $4$--$5\%$ slower than the same weights run at a fixed depth equal to their realized steps.
Prefill collapses to $1.00$--$1.40\times$, against $1.33$--$2.29\times$ at that fixed depth
(Table~\ref{tab:synchrony}). Appendix~\ref{app:feasibility} solves the cost model for $R$ from it.

\begin{table*}[!tp]

\begin{center}
\footnotesize
\setlength{\tabcolsep}{5pt}
\begin{tabular*}{\linewidth}{@{\extracolsep{\fill}}llccccc@{}}
\toprule
\multicolumn{1}{c}{\bf Model} & \multicolumn{1}{c}{\bf Workload}
& \multicolumn{1}{c}{$\varepsilon$} & \multicolumn{1}{c}{$\bar N$}
& \multicolumn{1}{c}{$\max_i N_i$} & \multicolumn{1}{c}{\bf Measured}
& \multicolumn{1}{c}{\bf Fixed at $\bar N$}
\\ \midrule
\multicolumn{7}{l}{\emph{370M, both models, nine repeats per cell}} \\
Mixture   & prefill, $b{=}1$  & 0.3 & 5.44 & 8 & $1.00\times$ & $1.33\times$ \\
Mixture   & prefill, $b{=}32$ & 0.6 & 2.17 & 5 & $1.40\times$ & $2.29\times$ \\
Mixture   & decode, $b{=}1$   & 0.5 & 2.86 & 5 & $1.93\times$ & $2.04\times$ \\
Mixture   & decode, $b{=}32$  & 0.5 & 2.90 & 5 & $1.65\times$ & $2.03\times$ \\[3pt]
Per-step, $\beta{=}0.1$ & prefill, $b{=}1$  & 0.7 & 2.93 & 6 & $1.18\times$ & $1.96\times$ \\
Per-step, $\beta{=}0.1$ & decode, $b{=}1$   & 0.7 & 2.59 & 5 & $2.08\times$ & $2.17\times$ \\
Per-step, $\beta{=}0.1$ & decode, $b{=}32$  & 0.7 & 2.48 & 6 & $1.62\times$ & $2.25\times$ \\[3pt]
\multicolumn{7}{l}{\emph{770M, per-step $\beta{=}0.1$, on a B200, 15 repeats per cell}} \\
Per-step, $\beta{=}0.1$ & prefill, $b{=}1$  & 0.7 & 2.77 & 6 & $1.17\times$ & $2.03\times$ \\
Per-step, $\beta{=}0.1$ & prefill, $b{=}32$ & 0.7 & 3.00 & 7 & $1.11\times$ & $1.96\times$ \\
Per-step, $\beta{=}0.1$ & decode, $b{=}1$   & 0.7 & 2.20 & 5 & $2.24\times$ & $2.35\times$ \\
Per-step, $\beta{=}0.1$ & decode, $b{=}32$  & 0.7 & 2.41 & 6 & $1.69\times$ & $2.28\times$ \\
\bottomrule
\end{tabular*}
\end{center}
\caption{Speedup against $T=8$. \emph{Measured}: timed with the exit rule. \emph{Fixed at $\bar N$}: the same weights run at a fixed
depth equal to their realized steps (interpolated between integer depths). $\bar N$ and $\max_i N_i$: the mean and the largest realized steps over the timed tokens.
The 370M block is timed on one H200 and the 770M block on the device named in its header; inputs are validation tokens.}
\label{tab:synchrony}
\end{table*}

\paragraph{Why the exit rule falls short of the fixed depth.} Table~\ref{tab:synchrony} measures how far the deepest token's steps, $\max_i N_i$, exceed the mean realized steps, $\bar N$, at 370M and 770M. It is the wait for the deepest token: among $2048$ positions
in one batched pass, some position reaches the cap even when the mean is $5.4$, and
that one position holds the whole pass \citep{dean2013tail}. \citet{del2023skipdecode} make each later position run no more layers than the one before it, so no later token can make the pass wait, for exactly this reason. Continuous batching \citep{yu2022orca,kwon2023efficient}
schedules at request granularity and does not remove that wait: the loop still runs
to the deepest active token in the pass. Gathering the active positions before every iteration,
as Mixture-of-Depths \citep{raposo2024mixture} does, escapes it. So does depth-wise batching \citep{bae2025relaxed,bae2025mixture},
which gives a stopped token's place to a token from another sequence instead of shrinking the batch;
\citet{bae2025mixture} measure up to $2.06$ times the throughput of a standard transformer this way, at a
larger batch that their smaller key-value cache allows and not counting the time to write keys and
values. We time only gathering. Timed on synthetic assignments of a depth to each position, with mean depth
$m \in \{2, 3, 4, 6\}$ at $65{,}536$ tokens per pass, at 140M and 770M, per-position gathering
reaches $0.91$--$0.99\times$ the throughput of running every position to $m$, while
hiding exited positions in place gains nothing ($0.99$--$1.00\times$ the throughput of running every position to the deepest position's depth): the pass still costs what its
deepest position does. So gathering nearly matches, but does not beat, a model run at the fixed depth $m$. The exit machinery adds a smaller cost of its own: at $\varepsilon = 0$, where no token stops early, running the exit rule is $1.03$--$1.08\times$ slower at batch $1$ than running the same depth without it,
amortizing to $1.001\times$ at batch $32$.

\subsection{The feasibility test: fits and consistency checks}
\label{app:feasibility}

This appendix checks the cost model of Section~\ref{sec:feasibility} against the timings of Appendix~\ref{app:timing}, and gives the numbers behind the test before training.

\paragraph{The layer-count bound.} Layer counts bound $f$ without any timing: treating every layer as
equal cost, $f \le n_r T/(n_p + n_r T + n_c)$, which is $0.80$ for our stacks ($n_p = n_r = n_c$) at
$T = 8$.

\paragraph{The per-row fit.} With $f$, $c(t)$ and $S_{\text{arch}}$ as in Section~\ref{sec:feasibility}, for each row of Table~\ref{tab:timing} we fit $f$ to that row's
$t{=}1$ speedup alone and predict its remaining five columns (Table~\ref{tab:timing}, last two columns).

Eight one-parameter fits, one per row, give forty predictions, each within the $0.070\times$
run-to-run variation of Table~\ref{tab:timing}, and the fit is the same on a second GPU type.

\paragraph{Why the fits fall below the bound.} The layer-count bound lies above all eight fitted
values by $0.01$ to $0.07$. The gap is cost outside the layers (embedding, unembedding, loss head),
which no layer count sees. Prefill $f$ climbs with scale while decode $f$ is flat, consistent with
that cost shrinking relative to layer compute as the model widens; a single $f$ fitted at 1.3B
mispredicts 140M prefill at $t{=}1$ by $0.46\times$.

\paragraph{The synchrony ratio.} Solving the cost model for $R$ on Table~\ref{tab:synchrony}, with each row's 370M $f$ from Table~\ref{tab:timing}, prefill or decode as the row's workload, gives $R$
against the number of tokens $B$ batched in a pass: $1.08$, $1.52$, $1.83$ and $2.32$ at $B = 1$, $32$,
$2048$ and $65{,}536$, pooled over seven rows differing in model and threshold ($0.3$--$0.7$), so $R$
describes a mix of models and thresholds, not a controlled curve in $B$; the per-step rows alone give $1.06$,
$1.66$ and $2.19$ at $B = 1$, $32$ and $2048$. The $R$ counted in the real decode is lower than the per-step rows' $1.66$ at $B = 32$ ($1.55$
at 370M, Table~\ref{tab:batcheddecode}): it is taken on the full split rather than the timed tokens, and
solving the cost model for $R$ from timings carries the cost model's residual (up to $0.056\times$, Table~\ref{tab:timing}).

\paragraph{The bound before training.} At 370M the bound of Section~\ref{sec:feasibility} splits into
the depth curve's drop from three iterations to the cap, $1.1079 - 1.0875 = 0.020$, plus the
largest lead at the cap among our models trained at the cap, $1.0875 - 1.0808 = 0.007$
(Appendix~\ref{app:earlyexit}, Table~\ref{tab:earlyexit}, $\varepsilon = 0$): about
$0.03$ BPB. With $R = 1.55$,
counted in the real decode, a pass for a head at three iterations runs $1.55 \times 3 = 4.65$ iterations; the curve there, interpolated between
$t{=}4$ and $5$, is $1.0930$, so the drop is $0.0055$ and the bound $0.012$.

\paragraph{The speedup at equal quality.} At 770M the per-step model's batched pass runs $4.37$ iterations
and leads the sampled-depth baseline there by $0.0030$ BPB (Appendix~\ref{app:decodematched}, Table~\ref{tab:batcheddecode}). Near that
depth the baseline's depth curve falls about $0.0049$ BPB per iteration ($1.0241$ at $t{=}4$, $1.0192$ at
$t{=}5$, from the full-split evaluation session of Appendix~\ref{app:fixedcontrol}; not tabulated), so it matches the model at about
$4.98$ iterations. Through the cost model at the decode $f = 0.792$ (Table~\ref{tab:timing}; fitted at one sequence, an approximation for $32$), that is
$c(4.98)/c(4.37) \approx 1.09$: at equal quality the model's pass delivers about $9\%$ more throughput
than the baseline's. Against the specialists it delivers none (Appendix~\ref{app:specialist}).

\paragraph{How steep would be enough.} At the depth each head's rule reaches, the depth curve would
have to fall by the head's BPB minus the cap's (Table~\ref{tab:steepness}): $7.6$ to $9.3$ times
the fall our curve actually shows, at all four scales. A steeper curve alone does not pass the test: Ouro's curve falls
$15.7$ times as far as ours, and what decides is how much of the fall lies below the depth a batched
pass runs (Appendix~\ref{app:ouro}). Retrained with gradient through all eight
iterations, the 370M sampled-depth baseline's curve falls $0.0219$ over $t{=}3$ to $8$ (the released baseline's span is $0.0204$), so it would still
have to be about $8.6$ times steeper.

\begin{table}[!tp]

\centering
\small
\setlength{\tabcolsep}{2pt}
\begin{tabular*}{\linewidth}{@{\extracolsep{\fill}}lccccc@{}}
\toprule
& & \bf Mixture & \bf Baseline & \bf Curve & \bf Would have \\
\bf Scale & \bf Steps & \bf BPB & \bf at the cap & \bf falls & \bf to fall \\ \midrule
140M & 3.020 & 1.4493 & 1.2159 & 0.0256 & $0.2334$ \\
370M & 3.017 & 1.2759 & 1.0875 & 0.0202 & $0.1884$ \\
770M & 2.885 & 1.1881 & 1.0157 & 0.0224 & $0.1724$ \\
1.3B & 2.925 & 1.1353 & 0.9693 & 0.0217 & $0.1660$ \\
\bottomrule
\end{tabular*}
\caption{How far the depth curve would have to fall for the quality condition to hold: the mixture-loss model's BPB at about three iterations ($\varepsilon{=}0.5$) minus the depth curve at the cap, beside the fall the curve actually shows from the head's steps to the cap. \emph{Steps}: realized steps at $\varepsilon{=}0.5$; the other columns are BPB.}
\label{tab:steepness}
\end{table}

\paragraph{Training at $T{=}32$: the depth curve is steeper, not flatter.} Two models at 140M on CC,
$2/2/2$, $11.2$B tokens, two seeds each (Table~\ref{tab:t32}): the sampled-depth baseline that
supplies the depth curve, and a per-step companion at $\beta = 0.1$. Comparing the same fraction of the cap, $t/T = 3/8$, the span $t{=}12 \to 32$ costs the sampled-depth baseline $0.0957$ and $0.0812$
BPB (its two seeds), against $0.0260$ for the same relative span at $T_{\text{train}}{=}8$ (Table~\ref{tab:pressure}, $1.2415 - 1.2155$), so there is more
quality to buy with depth, not less. In absolute terms the deeper cap reaches $1.2087$ at
$t{=}32$, $0.007$ to $0.011$ ahead of the $T_{\text{train}}{=}8$ baseline at its cap, for four times
the core's compute. Running below the training cap costs far more: at $t{=}8$ the two seeds
read $1.7466$ and $1.6305$, over $0.4$ BPB behind a model trained at $8$, consistent with a sampler
that rarely draws a shallow depth.
A deep cap is informative only if the run beats its own $t{=}8$ BPB there by more than the evaluation noise floor
(the \emph{deep-cap check}). The sampled-depth baseline passes by $0.54$ and $0.42$;
the per-step companion fails it ($0.0002$), and the cap-16 control below tests the per-step model.

\begin{table*}[!tp]

\begin{center}
\footnotesize
\setlength{\tabcolsep}{2.5pt}
\begin{tabular*}{\linewidth}{@{\extracolsep{\fill}}llcccccccccc@{}}
\toprule
& & \multicolumn{10}{c}{\bf forced depth $t$} \\
\multicolumn{1}{c}{\bf Model} & \multicolumn{1}{c}{\bf Seed}
& \multicolumn{1}{c}{$1$} & \multicolumn{1}{c}{$2$} & \multicolumn{1}{c}{$3$}
& \multicolumn{1}{c}{$4$} & \multicolumn{1}{c}{$6$} & \multicolumn{1}{c}{$8$}
& \multicolumn{1}{c}{$12$} & \multicolumn{1}{c}{$16$} & \multicolumn{1}{c}{$24$}
& \multicolumn{1}{c}{$32$} \\ \midrule
Sampled-depth & 42 & 7.2937 & 7.4076 & 5.0295 & 3.8793 & 2.7472 & 1.7466 & 1.3038 & 1.2311 & 1.2098 & 1.2081 \\
Sampled-depth & 7  & 5.9709 & 5.8090 & 4.6748 & 3.9846 & 2.5963 & 1.6305 & 1.2904 & 1.2285 & 1.2105 & 1.2092 \\[2pt]
Per-step, $\beta{=}0.1$ & 42 & 1.3337 & 1.2532 & 1.2290 & 1.2206 & 1.2163 & 1.2156 & 1.2154 & 1.2154 & 1.2154 & 1.2154 \\
Per-step, $\beta{=}0.1$ & 7  & 1.3386 & 1.2654 & 1.2430 & 1.2352 & 1.2313 & 1.2307 & 1.2305 & 1.2305 & 1.2305 & 1.2305 \\
\bottomrule
\end{tabular*}
\end{center}
\caption{Forced-depth BPB at $T_{\text{train}}{=}32$, 140M, CC, $2/2/2$, $11.2$B tokens,
weights frozen and only the forced depth $t$ changing. Two models, two seeds each; the anchor is $t{=}12$, at $t/T = 3/8$. \textbf{The baseline's cells below $t{=}8$ are not a usable depth curve: its training rarely drew such depths}.}
\label{tab:t32}
\end{table*}

\subsection{Gradient reach does not explain why the per-step model stops improving with depth: the cap-16 control}
\label{app:t16}

\paragraph{The comparison.} The per-step model, a sampled-depth baseline, and a specialist at
six iterations, about where the exit rule stops at $\varepsilon = 0.3$ ($6.5$ steps; Table~\ref{tab:t16exit}), all exist at $T{=}16$ (Table~\ref{tab:t16}). Under the deep-cap check at $t{=}16$ the baseline
passes ($0.0087$, $0.0090$) and the model fails ($0.0001$, $0.0002$). The per-step model sends gradients through all sixteen iterations and the baseline through eight
(at a cap of eight the baseline sends them through four, Table~\ref{tab:methods}), so the model with the longer gradient reach is the one that stops improving: after six
iterations it moves under $0.001$ over the remaining ten, where the baseline moves $0.024$. Against
a specialist trained at six iterations (three seeds, mean $1.2072$, s.d.\ $0.0004$) the
model's best forced-depth quality is worse by $0.0064$ and $0.0055$, both outside the evaluation noise floor. The model
is ahead of its baseline by more than the evaluation noise floor at every depth up to six and behind it from twelve on, by $0.0029$--$0.0068$ (outside the evaluation noise floor only for seed $42$), which is the
depth-robustness of Section~\ref{sec:results:recipe} at twice the cap. $\mathbb{E}[N]$ reaches
$5.56$ and $5.52$ of $16$ iterations, above the prior's mean of $3.84$.

\begin{table*}[!tp]

\begin{center}
\footnotesize
\setlength{\tabcolsep}{2.5pt}
\begin{tabular*}{\linewidth}{@{\extracolsep{\fill}}llccccccccc@{}}
\toprule
& & \multicolumn{9}{c}{\bf forced depth $t$} \\
\multicolumn{1}{c}{\bf Model} & \multicolumn{1}{c}{\bf Seed}
& \multicolumn{1}{c}{$1$} & \multicolumn{1}{c}{$2$} & \multicolumn{1}{c}{$3$}
& \multicolumn{1}{c}{$4$} & \multicolumn{1}{c}{$5$} & \multicolumn{1}{c}{$6$}
& \multicolumn{1}{c}{$8$} & \multicolumn{1}{c}{$12$} & \multicolumn{1}{c}{$16$} \\ \midrule
Sampled-depth & 42 & 1.7147 & 1.4347 & 1.3293 & 1.2763 & 1.2477 & 1.2311 & 1.2155 & 1.2078 & 1.2068 \\
Sampled-depth & 7  & 1.6856 & 1.4331 & 1.3310 & 1.2782 & 1.2496 & 1.2332 & 1.2177 & 1.2098 & 1.2087 \\[2pt]
Per-step, $\beta{=}0.1$ & 42 & 1.3324 & 1.2519 & 1.2274 & 1.2188 & 1.2156 & 1.2144 & 1.2137 & 1.2136 & 1.2136 \\
Per-step, $\beta{=}0.1$ & 7  & 1.3236 & 1.2490 & 1.2257 & 1.2176 & 1.2147 & 1.2136 & 1.2129 & 1.2127 & 1.2127 \\[2pt]
Specialist $T{=}6$ & 42 & --- & --- & --- & --- & --- & 1.2077 & --- & --- & --- \\
Specialist $T{=}6$ & 7  & --- & --- & --- & --- & --- & 1.2070 & --- & --- & --- \\
Specialist $T{=}6$ & 123 & --- & --- & --- & --- & --- & 1.2069 & --- & --- & --- \\[2pt]
$\mathbb{E}[N]$, per-step & 42 & 1.00 & 1.96 & 2.81 & 3.52 & 4.07 & 4.49 & 5.04 & 5.48 & 5.56 \\
$\mathbb{E}[N]$, per-step & 7  & 1.00 & 1.96 & 2.80 & 3.50 & 4.05 & 4.47 & 5.00 & 5.43 & 5.52 \\
\bottomrule
\end{tabular*}
\end{center}
\caption{Forced-depth BPB at $T_{\text{train}}{=}16$, 140M, CC, $2/2/2$, $11.2$B tokens, weights
frozen and only the forced depth $t$ changing. \textbf{The per-step model reaches all sixteen iterations
with gradient, the sampled-depth rows eight}. The specialist is shown at its trained depth. $\mathbb{E}[N]$: expected halting steps with the cap set to $t$.}
\label{tab:t16}
\end{table*}

\paragraph{The model at the depth its own rule realizes.} Sweeping the exit threshold at $T{=}16$
and reading out the exit state, the state at the iteration where a token stops, alone (Table~\ref{tab:t16exit}), the model never reaches the six-step
specialist: its best exit-rule result is $1.2128$ and $1.2119$, $+0.0056$ and $+0.0047$ against the six-step specialist's $1.2072$. Realized steps again exceed $\mathbb{E}[N]$
($6.5$ at $\varepsilon = 0.3$, against $\mathbb{E}[N] = 5.56$), and at $\varepsilon = 0$, the exit-state readout instead of the mixture changes BPB by $-0.0004$ and $-0.0003$, inside the evaluation noise floor.

\begin{table}[!t]

\begin{center}
\footnotesize
\setlength{\tabcolsep}{2.5pt}
\begin{tabular*}{\linewidth}{@{\extracolsep{\fill}}lccccc@{}}
\toprule
\multicolumn{1}{c}{$\varepsilon$} & \multicolumn{1}{c}{\bf Seed}
& \multicolumn{1}{c}{\bf Steps} & \multicolumn{1}{c}{\bf BPB}
& \multicolumn{1}{c}{\bf Specialist} & \multicolumn{1}{c}{\bf Model $-$ specialist} \\ \midrule
$0$    & 42 & 15.993 & 1.2132 & 1.2047 & $+0.0085$ \\
$0$    & 7  & 16.000 & 1.2124 & 1.2047 & $+0.0077$ \\[2pt]
$0.3$  & 42 & 6.531  & 1.2128 & 1.2065 & $+0.0063$ \\
$0.3$  & 7  & 6.491  & 1.2119 & 1.2066 & $+0.0053$ \\[2pt]
$0.5$  & 42 & 4.798  & 1.2139 & 1.2114 & $+0.0025$ \\
$0.5$  & 7  & 4.755  & 1.2130 & 1.2116 & $+0.0014$ \\[2pt]
$0.7$  & 42 & 3.567  & 1.2173 & 1.2188 & $-0.0015$ \\
$0.7$  & 7  & 3.518  & 1.2164 & 1.2193 & $-0.0029$ \\[2pt]
$0.9$  & 42 & 2.431  & 1.2292 & 1.2300 & $-0.0008$ \\
$0.9$  & 7  & 2.371  & 1.2283 & 1.2307 & $-0.0024$ \\[2pt]
$0.95$ & 42 & 2.093  & 1.2377 & 1.2338 & $+0.0039$ \\
$0.95$ & 7  & 2.035  & 1.2369 & 1.2345 & $+0.0024$ \\
\bottomrule
\end{tabular*}
\end{center}
\caption{Threshold early exit at $T_{\text{train}}{=}16$, 140M, CC, two seeds, exit-state readout. \emph{Steps}: realized steps, not $\mathbb{E}[N]$. \emph{Specialist}: the specialist's BPB at the row's realized steps, interpolated linearly between the
specialists of Appendix~\ref{app:specialist} (trained at $2$ to $5$ and $8$ iterations) and the six-iteration specialist trained at this cap,
and held at the eight-iteration value beyond eight. \textbf{At matched realized steps the two are inside the evaluation noise floor for
$\varepsilon \ge 0.5$}, the same no-advantage verdict as at a cap of eight.}
\label{tab:t16exit}
\end{table}

\section{Quality at depth}

\subsection{Quality at the training depth}
\label{app:main}

Table~\ref{tab:main} reports both corpora at the training depth $T=8$.

\begin{table*}[!tp]

\begin{center}
\small
\setlength{\tabcolsep}{4pt}
\begin{tabular*}{\linewidth}{@{\extracolsep{\fill}}llcccccc@{}}
\toprule
& & \multicolumn{3}{c}{\bf CC} & \multicolumn{3}{c}{\bf FW} \\
\multicolumn{1}{c}{\bf Scale} & \multicolumn{1}{c}{\bf Method}
& \multicolumn{1}{c}{$n$} & \multicolumn{1}{c}{\bf Core-Ext $\uparrow$} & \multicolumn{1}{c}{\bf BPB $\downarrow$}
& \multicolumn{1}{c}{$n$} & \multicolumn{1}{c}{\bf Core-Ext $\uparrow$} & \multicolumn{1}{c}{\bf BPB $\downarrow$}
\\ \midrule
140M & Dense                & 2 & 0.0488\,\tiny{$\pm$.0021} & 1.2576\,\tiny{$\pm$.0010} & -- & -- & -- \\
140M & Sampled-depth        & 3 & 0.0641\,\tiny{$\pm$.0010} & 1.2193\,\tiny{$\pm$.0026} & 3 & 0.0906\,\tiny{$\pm$.0111} & 0.9054\,\tiny{$\pm$.0012} \\
140M & ACT                  & 3 & \bf 0.0724\,\tiny{$\pm$.0021} & \bf 1.2027\,\tiny{$\pm$.0007} & -- & -- & -- \\
140M & Mixture  & 3 & 0.0676\,\tiny{$\pm$.0037} & 1.2052\,\tiny{$\pm$.0007} & 3 & \bf 0.1014\,\tiny{$\pm$.0051} & \bf 0.8983\,\tiny{$\pm$.0005} \\[2pt]
370M & Dense                & 2 & 0.0765\,\tiny{$\pm$.0027} & 1.1208\,\tiny{$\pm$.0011} & -- & -- & -- \\
370M & Sampled-depth        & 3/2$^{\ddagger}$ & 0.0851\,\tiny{$\pm$.0016} & 1.0875\,\tiny{$\pm$.0001}$^{\ddagger}$ & 3 & 0.1328\,\tiny{$\pm$.0008} & 0.8238\,\tiny{$\pm$.0005} \\
370M & ACT                  & 3 & 0.0867\,\tiny{$\pm$.0031} & \bf 1.0805\,\tiny{$\pm$.0004} & -- & -- & -- \\
370M & Mixture  & 3 & \bf 0.0918\,\tiny{$\pm$.0046} & \bf 1.0809\,\tiny{$\pm$.0003} & 3 & \bf 0.1437\,\tiny{$\pm$.0026} & \bf 0.8171\,\tiny{$\pm$.0002} \\[2pt]
770M & Dense                & 2 & 0.0913\,\tiny{$\pm$.0064} & 1.0423\,\tiny{$\pm$.0001} & -- & -- & -- \\
770M & Sampled-depth        & 3 & 0.0992\,\tiny{$\pm$.0023} & 1.0157\,\tiny{$\pm$.0003} & 3 & \bf 0.1587\,\tiny{$\pm$.0016} & 0.7788\,\tiny{$\pm$.0002} \\
770M & ACT                  & 3 & \bf 0.1053\,\tiny{$\pm$.0041} & \bf 1.0090\,\tiny{$\pm$.0003} & -- & -- & -- \\
770M & Mixture  & 3 & \bf 0.1053\,\tiny{$\pm$.0024} & \bf 1.0090\,\tiny{$\pm$.0012} & 3 & 0.1581\,\tiny{$\pm$.0035} & \bf 0.7718\,\tiny{$\pm$.0005} \\[2pt]
1.3B & Sampled-depth        & 3 & 0.1110\,\tiny{$\pm$.0062} & 0.9693\,\tiny{$\pm$.0010} & 3 & 0.1768\,\tiny{$\pm$.0035} & 0.7494\,\tiny{$\pm$.0007} \\
1.3B & ACT                  & 3 & \bf 0.1196\,\tiny{$\pm$.0076} & \bf 0.9643\,\tiny{$\pm$.0010} & -- & -- & -- \\
1.3B & Mixture  & 3 & 0.1169\,\tiny{$\pm$.0049} & \bf 0.9644\,\tiny{$\pm$.0017} & 3 & \bf 0.1792\,\tiny{$\pm$.0054} & \bf 0.7434\,\tiny{$\pm$.0009} \\
\bottomrule
\end{tabular*}
\end{center}
\caption{Quality at the training depth $T=8$. Core-Ext is the DCLM CORE-extended aggregate; BPB is measured on the validation
split of the training corpus. Mean $\pm$ s.d.\ over $n$ seeds, best per
column-block in bold, with any value within $0.0005$ of it. Dense models were trained on CC only, at 140M to 770M. Core-Ext
differences below $0.0042$ are unresolved. $^{\ddagger}$BPB over $n=2$. The mixture-loss BPB values differ from those of Tables~\ref{tab:decode} and~\ref{tab:earlyexit} by at most $0.0006$: a different evaluation session, within the evaluation noise floor.}
\label{tab:main}
\end{table*}

\textbf{The mixture-loss model beats the sampled-depth baseline} in all eight (scale, corpus) cells, by $0.005$ to
$0.014$ BPB (the exit-sweep session, Table~\ref{tab:earlyexit}, measures the baseline separately and gives $0.005$ to $0.010$ at $\varepsilon = 0$). The baseline is trained with the released recipe: it draws a random number of iterations per step and sends gradients through only the last four, whereas the halting models send them through all eight (gradient reach, Section~\ref{sec:setup:methods}). A model trained at a fixed eight iterations, with gradient through all eight, closes this gap at 140M (Appendix~\ref{app:fixedcontrol}), so the gap comes from the baseline's training schedule.

\textbf{On CC, looping beats a dense transformer at matched parameters}, and at 140M the mixture-loss model still beats
one trained with more compute. Trained on eight times the tokens ($89.6$B), more training compute than the looped model uses, the dense model reaches $1.2106$ BPB
($n{=}2$), better than sampled-depth ($1.2193$) and worse than the mixture-loss model ($1.2052$). The parameter-matched dense model runs three tenths of the layer computations per token ($n_p+n_r+n_c$ against $n_p+8n_r+n_c$).

\textbf{On CC, ACT matches the mixture-loss model at the training depth} from 370M to 1.3B (within $0.0004$ BPB in
three-seed means); at 140M ACT leads by $0.0025$.

\subsection{Specialists}
\label{app:specialist}
A specialist, a model trained to always run one fixed number of iterations with gradients through all of them, is a tighter reference than the sampled-depth baseline, because it is trained for exactly the depth it is compared at.

Table~\ref{tab:specialist} gives the specialists at 140M. The two depths an exit rule reaches, $T{=}3$ and $T{=}4$, carry three seeds
($1.2236 \pm .0006$ and $1.2152 \pm .0004$); the others are seed $42$. The ``about $0.02$ BPB'' of
Section~\ref{sec:results:main} is the $T{=}3$ specialist minus the $T{=}8$ specialist, $1.2236 - 1.2047 = 0.0189$.

\begin{table}[!t]

\begin{center}
\small
\setlength{\tabcolsep}{2.8pt}
\begin{tabular*}{\linewidth}{@{\extracolsep{\fill}}lccccc@{}}
\toprule
\multicolumn{1}{c}{\bf Model} & \multicolumn{1}{c}{$t{=}2$}
& \multicolumn{1}{c}{$t{=}3$} & \multicolumn{1}{c}{$t{=}4$}
& \multicolumn{1}{c}{$t{=}5$} & \multicolumn{1}{c}{$t{=}8$}
\\ \midrule
Specialist $T{=}2$ & \bf 1.2349 & 1.3124 & 1.3953 & 1.4353 & 1.4779 \\
Specialist $T{=}3$ & 1.6913 & \bf 1.2236 & 1.3212 & 1.3854 & 1.3434 \\
Specialist $T{=}4$ & 1.8764 & 1.7231 & \bf 1.2152 & 1.3088 & 1.3838 \\
Specialist $T{=}5$ & 1.8973 & 1.8333 & 1.6849 & \bf 1.2104 & 1.5053 \\
Specialist $T{=}8$ & 1.8785 & 1.8578 & 1.8347 & 1.7676 & \bf 1.2047 \\[4pt]
Per-step, $\beta{=}0.1$ & 1.2456 & 1.2251 & 1.2185 & 1.2164 & 1.2154 \\
Per-step, $\beta{=}1.0$ & 1.2344 & 1.2278 & 1.2261 & 1.2256 & 1.2255 \\
\bottomrule
\end{tabular*}
\end{center}
\caption{Specialists at 140M on CC, each trained at one number of iterations $T$ with gradients
through all of them. \textbf{Bold}: each specialist's lowest BPB, at its own training depth. Entries
off the diagonal are depths the model never saw in training. Entries are BPB with every token forced to run $t$ iterations; lower is better. Specialists at $T{=}3, 4$: three-seed means on the diagonal, seed $42$ elsewhere; other specialists: seed $42$. Per-step rows: $\beta{=}0.1$ seed $42$, $\beta{=}1.0$ three-seed means.}
\label{tab:specialist}
\end{table}
\paragraph{The per-step model against the specialists.} Per-step model ($\beta{=}0.1$, seed $42$) minus specialist at matched depth is
$+0.0107$, $+0.0015$, $+0.0033$, $+0.0060$ and $+0.0107$ at $T = 2,3,4,5,8$ at 140M. At 370M and 770M
all three seeds of each model were measured in one session, so the seed s.d.\ of the difference is the
criterion (Section~\ref{sec:setup:data}): $-0.0002 \pm .0003$ and $+0.0024 \pm .0010$
at $T = 3,5$ at 370M (specialists $1.0937$ and $1.0857$), and $-0.0002 \pm .0005$ and
$+0.0015 \pm .0004$ at 770M ($1.0196$ and $1.0128$; $\pm$: seed s.d.\ of the difference). Three 140M depths lose by
more than the evaluation noise floor and none wins by more, so at 140M the model saves no compute over a model trained
for the depth. At 370M and 770M the model ties the specialist at $T{=}3$ and trails it at $T{=}5$ by
more than twice that s.d., a resolved trail.

\paragraph{Specialists at the depth a batched pass pays.} Decoding $32$ sequences, a pass of the per-step
model runs $4.71$, $4.40$ and $4.37$ iterations at 140M, 370M and 770M (Table~\ref{tab:batcheddecode}).
Splitting passes between the specialists on either side of that depth ($T{=}4$ and $5$ at 140M, $T{=}3$
and $5$ above) runs them for the same average depth with no deepest token to wait for, at a BPB interpolated linearly
between the two: $1.2118$, $1.0881$ and $1.0149$ (the 140M $T{=}5$ specialist is seed $42$; the
others are three-seed means). The per-step model at $\varepsilon = 0.7$ reads
$1.2196$, $1.0908$ and $1.0171$ teacher-forced (Table~\ref{tab:adaptexit}), its best case, since in a pass each
position supplies keys and values only up to the depth that pass runs, not to the cap. It does not beat the specialists at
any scale: it trails by $0.0078$ at 140M, beyond the evaluation noise floor, and by $0.0027$ and $0.0022$ at 370M and
770M, inside it (the two sets were measured in different sessions). The comparison is matched on
compute, not memory: splitting passes between two specialists needs two sets of weights, which is where the head keeps its advantage.

\subsection{A sampled-depth baseline trained at shallower depths}
\label{app:shallow}
The released sampled-depth baseline draws each training sequence's depth from a Poisson distribution with mean $8$, so only about $4\%$ of its sequences run three or fewer iterations, the depths the per-step model's exit rule reaches. We retrain it at 140M on CC with the same recipe and mean $4$ (about $43\%$ of sequences at three or fewer), and with mean $2$, which under the same recipe also backpropagates through two iterations instead of four. Three seeds each; all models in Table~\ref{tab:shallow} are evaluated in one session, so differences are read against the seed s.d.\ of the difference. At three iterations the mean-$4$ baseline is $1.2264$, within $0.0029 \pm 0.0010$ of the $T{=}3$ specialist. Against it, the per-step model at $\varepsilon = 0.7$ leads, teacher-forced, by $0.0064 \pm 0.0012$ at its realized steps ($3.11$), against $0.0205 \pm 0.0013$ over the released baseline. The exit rule's allocation, measured against a uniform depth of equal compute, is worth $0.0048$ against either (the model's own curve at uniform depth, $1.2244$, against its exit state, $1.2196$), so depth-robustness is $0.0157$ against the released baseline and $0.0016$ against the mean-$4$ one, inside its s.d.\ (about $0.0014$). Shuffled depths, the measure of Section~\ref{sec:results:lead}, give allocation a larger share (Appendix~\ref{app:fixedcontrol}). At one iteration the per-step model is $0.036$ BPB worse than the mean-$4$ baseline. At the $4.71$ iterations a batched pass over $32$ sequences pays (teacher-forced in this session), it leads the released baseline by $0.0028 \pm 0.0010$, beyond twice the s.d., and the mean-$4$ baseline by $0.0021 \pm 0.0012$, inside it; both are teacher-forced, which favors the model (Section~\ref{sec:bg:props}), and the reading of Section~\ref{sec:batching} does not change. The mean-$2$ baseline is worse than the mean-$4$ one at every depth from two on.

\begin{table}[!t]
\centering
\scriptsize
\setlength{\tabcolsep}{1.5pt}
\begin{tabular*}{\linewidth}{@{\extracolsep{\fill}}lccccc@{}}
\toprule
$t$ & \bf Released & \bf Mean $4$ & \bf Mean $2$ & \bf \begin{tabular}[b]{@{}c@{}}Specialist\\ $T{=}3$\end{tabular} & \bf Per-step \\ \midrule
1 & 1.3801\,\tiny{$\pm$.0015} & 1.2808\,\tiny{$\pm$.0006} & 1.2574\,\tiny{$\pm$.0015} & 1.8779\,\tiny{$\pm$.060} & 1.3166\,\tiny{$\pm$.0044} \\
2 & 1.2779\,\tiny{$\pm$.0006} & 1.2388\,\tiny{$\pm$.0005} & 1.2409\,\tiny{$\pm$.0008} & 1.7251\,\tiny{$\pm$.038} & 1.2457\,\tiny{$\pm$.0019} \\
3 & 1.2418\,\tiny{$\pm$.0007} & 1.2264\,\tiny{$\pm$.0007} & 1.2375\,\tiny{$\pm$.0007} & 1.2236\,\tiny{$\pm$.0007} & 1.2251\,\tiny{$\pm$.0012} \\
4 & 1.2269\,\tiny{$\pm$.0006} & 1.2225\,\tiny{$\pm$.0008} & 1.2367\,\tiny{$\pm$.0006} & 1.3167\,\tiny{$\pm$.0063} & 1.2185\,\tiny{$\pm$.0010} \\
8 & 1.2160\,\tiny{$\pm$.0006} & 1.2208\,\tiny{$\pm$.0008} & 1.2364\,\tiny{$\pm$.0006} & 1.3985\,\tiny{$\pm$.103} & 1.2155\,\tiny{$\pm$.0008} \\
\bottomrule
\end{tabular*}
\caption{\textbf{Sampled-depth baselines trained at shallower depths.} 140M, CC, three seeds each, one evaluation session. BPB with every token forced to run $t$ iterations. \emph{Released}: Poisson mean $8$, the baseline of the main text. Per-step: $\beta{=}0.1$. $\pm$: seed s.d.; lower is better.}
\label{tab:shallow}
\end{table}

\section{Realized compute under an exit rule}

\subsection{Realized compute under early exit, all thresholds}
\label{app:earlyexit}
At every scale, the mixture-loss model's full-depth lead lasts only to $7.0$--$7.5$ realized steps.

Table~\ref{tab:earlyexit} gives the thresholds plotted in Figure~\ref{fig:exitcurve} (370M) and the same sweep at
140M, where the baseline has three seeds of its own, and at 770M and 1.3B.

\begin{table}[!t]

\begin{center}
\small
\setlength{\tabcolsep}{2pt}
\begin{tabular*}{\linewidth}{@{\extracolsep{\fill}}llccccc@{}}
\toprule
\multicolumn{1}{c}{\bf Scale} & \multicolumn{1}{c}{$\varepsilon$}
& \multicolumn{1}{c}{\bf Steps $\downarrow$} & \multicolumn{1}{c}{\bf Mixture $\downarrow$}
& \multicolumn{1}{c}{\bf Baseline $\downarrow$} & \multicolumn{1}{c}{$\Delta$ $\downarrow$}
\\ \midrule
140M & 0.0  & 8.000 & 1.2058\,\tiny{$\pm$.0006} & 1.2159 & $-0.0101$ \\
140M & 0.1  & 7.860 & 1.2067\,\tiny{$\pm$.0011} & 1.2160 & $-0.0093$ \\
140M & 0.2  & 6.993 & 1.2301\,\tiny{$\pm$.0080} & 1.2169 & $+0.0132$ \\
140M & 0.3  & 5.755 & 1.2855\,\tiny{$\pm$.0148} & 1.2185 & $+0.0670$ \\
140M & 0.5  & 3.020 & 1.4493\,\tiny{$\pm$.0165} & 1.2415 & $+0.2078$ \\[2pt]
370M & 0.0  & 8.000 & 1.0808 & 1.0875 & $-0.0067$ \\
370M & 0.1  & 7.762 & 1.0814 & 1.0877 & $-0.0063$ \\
370M & 0.2  & 6.688 & 1.0971 & 1.0885 & $+0.0086$ \\
370M & 0.3  & 5.466 & 1.1375 & 1.0902 & $+0.0473$ \\
370M & 0.5  & 3.017 & 1.2759 & 1.1077 & $+0.1682$ \\[2pt]
770M & 0.0  & 8.000 & 1.0090 & 1.0157 & $-0.0067$ \\
770M & 0.1  & 7.671 & 1.0094 & 1.0159 & $-0.0065$ \\
770M & 0.2  & 6.492 & 1.0221 & 1.0167 & $+0.0054$ \\
770M & 0.3  & 5.258 & 1.0558 & 1.0186 & $+0.0372$ \\
770M & 0.5  & 2.885 & 1.1881 & 1.0381 & $+0.1500$ \\[2pt]
1.3B & 0.0  & 8.000 & 0.9644 & 0.9693 & $-0.0049$ \\
1.3B & 0.1  & 7.611 & 0.9648 & 0.9696 & $-0.0048$ \\
1.3B & 0.2  & 6.381 & 0.9784 & 0.9705 & $+0.0079$ \\
1.3B & 0.3  & 5.179 & 1.0127 & 0.9727 & $+0.0400$ \\
1.3B & 0.5  & 2.925 & 1.1353 & 0.9910 & $+0.1443$ \\
\bottomrule
\end{tabular*}
\end{center}
\caption{Realized compute under threshold early exit, CC. The mixture-loss column is a three-seed mean; the Baseline column
uses three seeds at 770M, seed $42$ at 370M, and three seeds at $t = 1, 3, 5, 6$ at 1.3B. Larger $\varepsilon$ exits
earlier. ``Baseline'' is the sampled-depth baseline at the same number of iterations, interpolated
between integer $T$ on that scale's depth curve; the curve is convex, so interpolation
favors the mixture-loss model. $\varepsilon = 0$ runs all eight iterations, so its $\Delta$ is the model's margin at full depth.
At 140M the baseline has three seeds of its own, the mixture-loss model uses the mixture readout, and $\pm$ is the seed s.d. All values are BPB; lower is better. $\Delta$ is the mixture-loss model minus the baseline, so a negative value favors the model.}
\label{tab:earlyexit}
\end{table}

\paragraph{What the prior alone implies.} The prior is geometric with rate $1/\kappa = 1/4$ on
$\{1,\dots,8\}$, so the halting distribution that minimizes the divergence to it is
$p_t \propto 0.25 \cdot 0.75^{t-1}$, with mean $3.11$ and running total of probability $0.848$, $0.913$ and
$0.963$ at $t = 5$, $6$ and $7$. A head that matched it exactly would therefore reach $90\%$ of its
mass at step $6$. At that threshold the learned heads run $7.6$ to $7.9$ steps
(Table~\ref{tab:earlyexit}), so their late mass is heavier than the prior's and
is not inherited from it.

\paragraph{What a batched pass pays at $\varepsilon = 0.1$.} Table~\ref{tab:chain}'s $R\,\bar N = 8.0$ for the 370M mixture-loss model is derived, not timed. Its tokens realize $7.762$ of $8$ steps, and a token that stops early runs at most seven, so at least $76\%$ of tokens run all eight iterations. A pass over $32$ sequences stops before the cap only if all $32$ of its tokens are among the other $24\%$, which, if tokens stop independently, happens with probability below $0.24^{32}$. Every pass therefore runs eight iterations. Without the independence assumption, the pass still runs at least $\bar N = 7.76$.

The crossover, below which the model is behind the baseline, falls at about $7.5$ realized steps at 140M. The model's
seed s.d.\ grows $27$-fold, from $0.0006$ at full depth to $0.0165$ at $\varepsilon{=}0.5$.

\paragraph{What is read out at each threshold.} The rule fixes \emph{when} a token stops; the state
emitted is a separate choice. The \emph{mixture} readout emits the mixture of the states computed so far, each keeping its weight, with the probability not yet used placed on the state at the exit, the construction of
\citet{tan2023sparse}, and \emph{exit-state} readout emits that step's state alone, as PonderNet's
test-time rule does. Each model is read out as its loss trained it, the mixture-loss model with the mixture
and the per-step model with the exit state, the choice that favors each model; Appendix~\ref{app:sampled}
measures both readouts for the mixture-loss model.

\paragraph{The per-step model read as a shortened mixture.} To separate
the loss from the readout, Table~\ref{tab:w1mixture} reads the per-step $\beta{=}0.1$ model both
ways, three seeds per scale, against the sampled-depth baseline. Under the shortened mixture (the mixture of the states up to the exit) the model stays ahead at every scale and threshold,
within $0.0011$ BPB of its exit-state margin.

\begin{table}[!t]
\centering
\scriptsize
\setlength{\tabcolsep}{2pt}
\begin{tabular*}{\linewidth}{@{\extracolsep{\fill}}lccccc@{}}
\toprule
\bf Scale & $\varepsilon$ & $\bar N$ & \bf Mixture & \bf Exit state & \bf Over own \\ \midrule
140M & 0.5  & 4.08 & $-0.0095$ & $-0.0106$ & $-0.0025$ \\
     & 0.7  & 3.11 & $-0.0199$ & $-0.0205$ & $-0.0048$ \\
     & 0.95 & 1.88 & $-0.0503$ & $-0.0504$ & $-0.0143$ \\[2pt]
370M & 0.5  & 3.79 & $-0.0100$ & $-0.0108$ & $-0.0025$ \\
     & 0.7  & 2.83 & $-0.0214$ & $-0.0219$ & $-0.0050$ \\
     & 0.95 & 1.63 & $-0.0589$ & $-0.0590$ & $-0.0174$ \\[2pt]
770M & 0.5  & 3.72 & $-0.0123$ & $-0.0131$ & $-0.0027$ \\
     & 0.7  & 2.75 & $-0.0241$ & $-0.0246$ & $-0.0055$ \\
     & 0.95 & 1.56 & $-0.0623$ & $-0.0623$ & $-0.0184$ \\
\bottomrule
\end{tabular*}
\caption{Per-step $\beta{=}0.1$ model, both readouts, CC. \emph{Mixture},
\emph{Exit state}: model minus sampled-depth baseline at $\bar N$. \emph{Over own}: the exit-state margin
minus the margin that the model's own forced-depth curve gives at the same $\bar N$ (every token run at that depth). Each model and each baseline uses three seeds.
Seed s.d.\ of each margin at most $0.0011$, except $0.0027$ at 140M, $\varepsilon{=}0.95$. At 140M and 370M the baseline's three seeds were measured in the same session as the model, so margins differ slightly from Table~\ref{tab:adaptexit}.}
\label{tab:w1mixture}
\end{table}

\subsection{Threshold early exit on the per-step model under a strong prior}
\label{app:adaptexit}
With the per-step loss and a strong prior the rule does stop tokens early, and what it wins is depth-robustness more than allocation.

\begin{table*}[!t]

\begin{center}
\footnotesize
\setlength{\tabcolsep}{4.5pt}
\begin{tabular*}{\linewidth}{@{\extracolsep{\fill}}llccccl@{}}
\toprule
\multicolumn{1}{c}{\bf Model} & \multicolumn{1}{c}{$\varepsilon$}
& \multicolumn{1}{c}{\bf Steps $\downarrow$} & \multicolumn{1}{c}{\bf BPB $\downarrow$}
& \multicolumn{1}{c}{\bf Baseline} & \multicolumn{1}{c}{$\Delta$ $\downarrow$}
& \multicolumn{1}{c}{\bf Depth equivalent / steps [floor interval]}
\\ \midrule
\multicolumn{7}{l}{\emph{140M; $\beta{=}0.1$ rows three seeds, $\beta{=}1.0$ rows seed $42$}} \\
$\beta{=}0.1$ & 0.0  & 7.997 & 1.2143 & 1.2155 & $-0.0012$ & -- \\
$\beta{=}0.1$ & 0.5  & 4.074 & 1.2159 & 1.2261 & $-0.0102$ & $1.86\times$ [$1.22$, open] \\
$\beta{=}0.1$ & 0.7  & 3.113 & 1.2196 & 1.2398 & $-0.0202$ & $1.67\times$ [$1.41$, open] \\
$\beta{=}0.1$ & 0.9  & 2.160 & 1.2317 & 1.2721 & $-0.0404$ & $1.69\times$ [$1.56$, $1.83$] \\
$\beta{=}0.1$ & 0.95 & 1.879 & 1.2399 & 1.2905 & $-0.0506$ & $1.65\times$ [$1.56$, $1.81$] \\[2pt]
$\beta{=}1.0$ & 0.3  & 4.003 & 1.2233 & 1.2309 & $-0.0076$ & $1.25\times$ [$1.10$, $1.67$] \\
$\beta{=}1.0$ & 0.5  & 2.999 & 1.2257 & 1.2478 & $-0.0221$ & $1.56\times$ [$1.37$, $1.85$] \\
$\beta{=}1.0$ & 0.6  & 2.011 & 1.2325 & 1.2877 & $-0.0552$ & $1.94\times$ [$1.81$, $2.17$] \\[2pt]
\multicolumn{7}{l}{\emph{140M, mixture loss at $\beta{=}0.1$: the prior without the per-step loss, three seeds}} \\
Mixture, $\beta{=}0.1$ & 0.0  & 8.000 & 1.2062\,\tiny{$\pm$.0010} & 1.2159 & $-0.0097$ & -- \\
Mixture, $\beta{=}0.1$ & 0.1  & 6.813 & 1.2235\,\tiny{$\pm$.0018} & 1.2170 & $+0.0064$ & -- \\
Mixture, $\beta{=}0.1$ & 0.3  & 4.038 & 1.3523\,\tiny{$\pm$.0120} & 1.2267 & $+0.1257$ & -- \\
Mixture, $\beta{=}0.1$ & 0.5  & 2.937 & 1.4301\,\tiny{$\pm$.0125} & 1.2441 & $+0.1860$ & -- \\
Mixture, $\beta{=}0.1$ & 0.7  & 1.996 & 1.5485\,\tiny{$\pm$.0207} & 1.2783 & $+0.2702$ & -- \\
Mixture, $\beta{=}0.1$ & 0.9  & 1.000 & 1.8507\,\tiny{$\pm$.0439} & 1.3801 & $+0.4706$ & -- \\[4pt]
\multicolumn{7}{l}{\emph{370M, $\beta = 0.1$, three seeds}} \\
$\beta{=}0.1$ & 0.0  & 7.997 & 1.0864 & 1.0875 & $-0.0011$ & -- \\
$\beta{=}0.1$ & 0.3  & 4.977 & 1.0867 & 1.0913 & $-0.0046$ & $>1.61\times$ \\
$\beta{=}0.1$ & 0.5  & 3.788 & 1.0877 & 1.0987 & $-0.0110$ & $2.03\times$ [$1.27$, open] \\
$\beta{=}0.1$ & 0.7  & 2.820 & 1.0908 & 1.1131 & $-0.0223$ & $1.83\times$ [$1.49$, open] \\
$\beta{=}0.1$ & 0.95 & 1.620 & 1.1086 & 1.1689 & $-0.0603$ & $1.84\times$ [$1.74$, $2.05$] \\[4pt]
\multicolumn{7}{l}{\emph{770M, $\beta = 0.1$, three seeds}} \\
$\beta{=}0.1$ & 0.0  & 7.997 & 1.0128 & 1.0157 & $-0.0029$ & -- \\
$\beta{=}0.1$ & 0.3  & 4.903 & 1.0130 & 1.0197 & $-0.0067$ & $>1.63\times$ \\
$\beta{=}0.1$ & 0.5  & 3.720 & 1.0140 & 1.0271 & $-0.0131$ & $>2.15\times$ \\
$\beta{=}0.1$ & 0.7  & 2.752 & 1.0171 & 1.0417 & $-0.0246$ & $2.17\times$ [$1.64$, open] \\
$\beta{=}0.1$ & 0.9  & 1.786 & 1.0276 & 1.0789 & $-0.0513$ & $2.06\times$ [$1.84$, $2.34$] \\
$\beta{=}0.1$ & 0.95 & 1.564 & 1.0341 & 1.0965 & $-0.0624$ & $1.97\times$ [$1.84$, $2.22$] \\
\bottomrule
\end{tabular*}
\end{center}
\caption{Threshold early exit on the per-step model under a strong prior, CC; larger $\varepsilon$ exits earlier.
``Baseline'' is the sampled-depth baseline interpolated to the same realized steps. The depth equivalent is the depth that baseline needs to match
the model's BPB, so $1.86\times$ means the baseline needs $1.86$ times as many iterations as the model runs. \textbf{The bracketed interval recomputes it after moving the model's BPB by the evaluation noise floor and is the figure to read;} it includes only the evaluation noise floor, so the true uncertainty is at least this large. ``Open'' intervals have no upper end (see below the table). The mixture-loss rows are the prior without the per-step loss, against the sampled-depth baseline's three-seed curve; $\pm$ is the seed s.d.}
\label{tab:adaptexit}
\end{table*}

\paragraph{Reading Table~\ref{tab:adaptexit}.} The $\Delta$ column, model minus the sampled-depth
baseline at the same realized steps, is read directly. The depth-equivalent column
reads the baseline's curve backwards, and near full depth that curve falls by only $0.00075$--$0.0013$ BPB per step, so a floor-sized shift moves
the result a lot. The bracketed intervals include only the evaluation noise floor, so the true uncertainty is at least this large; they are not confidence intervals. An
interval open above, or a value written $>$, means a floor-sized improvement already puts the model
below the baseline's full-depth value, so no crossing exists. Model rows' seed s.d.\ is at most $0.0009$ BPB. At $\beta = 1.0$ every realized-step value sits within $0.011$ of an integer, so the strong
prior has turned the head into the same shallow depth for every token.

\paragraph{The lead, split into two parts.} Reading the model's own forced-depth curve at the realized
steps splits each $\Delta$ in two (Table~\ref{tab:w1mixture}, three seeds). Giving every token the same
depth, equal to the realized steps, removes $21$ to $30\%$ of the lead ($0.0025$ to $0.0184$ BPB;
about a quarter at $\varepsilon \le 0.7$), so that share is what the exit rule's allocation adds, measured against a uniform depth of equal compute (shuffled depths, the measure of Section~\ref{sec:results:lead}, give a larger share; Appendix~\ref{app:fixedcontrol}); the rest is depth-robustness, a property of how the model
was trained and not obtainable by running the baseline shallower. At 140M, the specialists interpolated to about $2.8$
iterations match the $\beta{=}1.0$ model (seed $42$) at $3$ realized steps (Appendix~\ref{app:specialist}).

\paragraph{At 770M the per-step model is also ahead at every forced depth measured.} Three seeds a side,
the model leads the sampled-depth baseline at forced
$t = 1$--$6$ and $8$ by $0.0621$, $0.0297$, $0.0155$, $0.0084$, $0.0048$, $0.0030$ and $0.0018$ BPB,
each at least eight standard errors of the difference. At forced full depth the lead ($0.0018$) is about what the 770M baseline gains when it is also trained with gradient through all eight iterations ($0.0017$; Appendix~\ref{app:fixedcontrol}).

\paragraph{The per-step loss is needed: a test under an exit rule.} With the prior but the mixture loss, the rule stops tokens early and quality
collapses (Table~\ref{tab:adaptexit}, mixture-loss rows).

\noindent At two realized steps the mixture-loss model is $0.270$ BPB worse than the sampled-depth baseline run to that depth, where
the per-step model with the strong prior at $1.88$ steps is $0.051$ better: opposite signs at nearby depth, three seeds
for each model. So the finding that both the per-step loss and the strong prior are needed rests on a sign difference, not only on the size of the differences in
Table~\ref{tab:twobytwo}.

\paragraph{The rule and the depth curve.} Every row uses the
exit rule of Section~\ref{sec:bg:props}; a per-step gate that stops a token when $\lambda_{i,t} > \theta$ on the same $\beta{=}0.1$ weights
traces the same curve to within $0.0009$ BPB at matched realized steps. The \emph{Baseline} column is
interpolated on the depth curve: at 140M
seed-$42$ curves (Appendix~\ref{app:pressure}'s for the $\beta{=}0.1$ rows, a second for the $\beta{=}1.0$
rows), at 370M a seed-$42$ curve ($1.2214$, $1.1368$, $1.1079$,
$1.0962$, $1.0912$, $1.0890$, $1.0875$ at $t = 1$--$6$ and $8$, seeds agreeing to within $0.0004$ past
$t{=}1$), and at 770M the three-seed depth curve of the forced-depth paragraph.

\subsection{Section~\ref{sec:feasibility}'s test applied to a released checkpoint}
\label{app:ouro}
\citet{zhu2025scaling}'s
Ouro is the only released model we know of that has a trained halting head (which Ouro calls its exit gate), an exit rule its released code implements, and loops over all of its layers (no prelude or coda). \citet{geiping2025scaling} train no halting head and already report
realized steps; \citet{tan2023sparse}'s mixture-loss head is below our smallest scale.

\paragraph{The rule we sweep is the rule they ship.} Their released \texttt{modeling\_ouro.py}
thresholds the running total of the halting probability and force-halts at the cap, Section~\ref{sec:bg:props}'s rule;
it has no per-step gate that stops a token when $\lambda_{i,t} > \theta$. Both of Ouro's models loop the full stack
($n_p = n_c = 0$), so $f$ is close to $1$ and the speedup is at most four, Ouro's cap. Their forward runs every
loop step for every token and selects the exit state afterwards, so its quality figures, like our
teacher-forced ones, keep every position's deeper keys and values
(Appendix~\ref{app:decodematched} measures the single-token case).

\paragraph{What the head reports, and what a pass runs.} The released gate of Ouro's 2.6B model gives
$\mathbb{E}[N] = 3.410$ of $4$. Sweeping Ouro's exit threshold $\tau_{\text{Ouro}}$ from $0.05$ to $1.0$ (a token stops once its running total of stopping probability reaches it, so a smaller value exits earlier), realized
steps run from $2.444$ to $4.000$, but a pass over $32$ tokens runs $3.61$ to $4.00$ iterations and a
pass over $2048$ tokens $3.98$ or more at every setting. Where quality holds the rule barely stops tokens early: at
$\tau_{\text{Ouro}} = 0.5$, within $0.0005$ BPB of full depth, it realizes $3.550$ of $4$ steps and a $32$-token pass
runs $3.96$; at $\tau_{\text{Ouro}} = 0.9$ it realizes $3.998$. Ouro's 1.4B model behaves the
same ($\mathbb{E}[N] = 3.502$; $2.669$ realized at $\tau_{\text{Ouro}} = 0.05$, and a pass runs $3.71$ with $32$ tokens and $4.000$ with $2048$). With
$\mathbb{E}[N]$ well below the cap, what keeps the rule from stopping tokens early is the shape of the halting
distribution, not the cap. Emitting one iteration instead of the mixture costs $0.0017$ BPB,
against $0.17$--$0.20$ for our mixture-loss models, as Section~\ref{sec:results:perstep} predicts for
a per-step loss.

\paragraph{The gate allocates well.} Table~\ref{tab:ourobuckets} gives per-token loss at every fixed
depth, bucketed by the step the gate chose. In each bucket past one iteration, the last iteration the gate takes
is worth $1.14$--$1.67$ nats of loss per token and the first it declines under $0.05$: the gate stops within one
iteration of where returns collapse. Against the same mean depth assigned without regard to the token ($55.5\%$ at two iterations,
$44.5\%$ at three), the allocation is worth $0.30$ BPB, seventy times the evaluation noise floor (the same comparison gives $0.28$ at
1.4B).

\begin{table}[!t]
\centering
\footnotesize
\setlength{\tabcolsep}{3pt}
\begin{tabular*}{\linewidth}{@{\extracolsep{\fill}}lccccc@{}}
\toprule
Gate chose & Share & $t{=}1$ & $t{=}2$ & $t{=}3$ & $t{=}4$ \\ \midrule
$1$ & $18.5\%$ & \bf 0.6665 & 0.6677 & 0.6613 & 0.6601 \\
$2$ & $30.6\%$ & 4.0175 & \bf 2.3504 & 2.3113 & 2.3033 \\
$3$ & $38.8\%$ & 7.2121 & 4.9394 & \bf 3.3773 & 3.3300 \\
$4$ & $12.1\%$ & 8.3433 & 7.4025 & 5.0786 & \bf 3.9426 \\ \midrule
Corpus BPB & & 2.4907 & 1.7641 & 1.3296 & 1.2531 \\
\bottomrule
\end{tabular*}
\caption{Ouro 2.6B at $\tau_{\text{Ouro}} = 0.05$: mean per-token loss (nats) at every fixed readout depth $t$,
bucketed by the step the gate chose (bold).}
\label{tab:ourobuckets}
\end{table}

\paragraph{The test on a curve over fifteen times steeper.} Ouro has no sampled-depth baseline, so its depth curve is Ouro itself run at each fixed depth (Table~\ref{tab:ourobuckets}, last row). It falls $0.318$ BPB over the
band its gate reaches ($2.44$ to $4.00$ steps), against $0.020$ over ours ($t{=}3$ to $8$ at 370M). At $\tau_{\text{Ouro}} = 0.05$ the
gate scores $1.2707$ BPB at $2.444$ realized steps. Read against the depth curve at the depth a pass of each
width runs, it wins by $0.300$ at one token per pass ($1.5710$ static), by $0.0124$ at $32$
tokens ($1.2831$ at $3.608$), and loses by $0.0163$ at $2048$ ($1.2544$ at $3.983$). So the quality
condition holds at one and $32$ tokens per pass, teacher-forced, and fails at $2048$ tokens per pass, where the pass
runs $3.98$ of $4$ iterations, a speedup of $1.004\times$. Two biases favor the gate: the curve is convex, so the straight line between $t{=}3$
and $4$ overstates the baseline at $3.608$; and the baseline is one set of weights truncated, not a
model trained for depth $3.6$. The $0.0124$ is an upper bound on both counts. Noise on this
checkpoint is at most $0.0015$ BPB.

\paragraph{What sets the depth curve's slope.} Raising the training cap steepens it
(Appendix~\ref{app:feasibility}), though those runs also differ in gradient reach and in how training depths are drawn; raising the
baseline's gradient reach alone moves its depth curve's drop from $t{=}3$ to $8$ by $0.0015$ at 370M, inside the evaluation noise floor, and by $0.0010$ at
770M (Appendices~\ref{app:feasibility} and~\ref{app:fixedcontrol}). Flatness is therefore
not intrinsic to the architecture.

\subsection{Pricing the exit rule for a batched pass}
\label{app:priced}
Pricing a pass means counting the iterations it runs, $R\,\bar N$ (Section~\ref{sec:bg:props}), instead of the average, and comparing at that depth.

\paragraph{Which $R$.} Section~\ref{sec:feasibility}'s condition reads the sampled-depth baseline at
$R\,\bar N$, the depth a batched pass runs. For a pass over $32$ sequences (one token each), a real decode counts $R$ directly (Appendix~\ref{app:decodematched}):
$1.51$--$1.58$ at $\varepsilon{=}0.7$ and $1.74$--$1.93$ at $0.95$, where the mean falls faster than the
deepest token. At the other widths $R$ depends on the depth distribution's
shape, so we take the $R$ that Appendix~\ref{app:feasibility} recovers from the per-step $\beta{=}0.1$ rows of Table~\ref{tab:synchrony} at
$\varepsilon{=}0.7$: $1.06$ and $2.19$ at $1$ and $2048$ tokens per
pass. For 370M at $\varepsilon{=}0.7$ this is the same model at the same threshold; the other blocks carry it over, as their headers state. At 770M, $R$ is measured on its own model ($1.10$, $2.35$ and $2.34$). At 140M and 370M the
$65{,}536$-token entry borrows a mixture-loss row ($\dagger$).

\begin{figure}[!t]
\begin{center}
\includegraphics[width=\columnwidth]{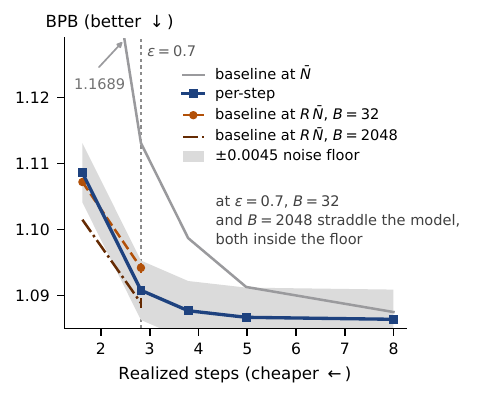}
\end{center}
\caption{\textbf{The per-step loss with a strong prior, and what pricing the pass does to it.} 370M, CC; model three seeds, baseline seed $42$. Against the sampled-depth baseline
read at $\bar N$ the model wins everywhere. Read at the depth a batched pass runs,
$R\,\bar N$, the baseline improves as the pass widens until it straddles the model, and
further left, the baseline at the widest pass is ahead outright. Legend: the baseline read at $R\,\bar N$ for a pass of $32$ tokens
(decoding $32$ sequences; $R$ counted in a real decode) and of $2048$ tokens (one prompt; $R$ from the timing table).
Here ``static'' in the legend is the sampled-depth baseline. Dashed line: $\varepsilon = 0.7$, the threshold used. The point at $1.1689$ lies above the axis. Model BPB is teacher-forced. The band is the evaluation noise floor, either side.}
\label{fig:priced}
\end{figure}

\paragraph{What pricing leaves (Figure~\ref{fig:priced}, Table~\ref{tab:priced}).} At one token per
pass the model stays ahead at all three scales, teacher-forced. Prefill is inside the evaluation noise floor at every scale at
$\varepsilon{=}0.7$; at 370M and $\varepsilon{=}0.95$ it favors the sampled-depth baseline by $0.0071$ and $0.0096$, and the second
exceeds twice the evaluation noise floor. The exit rule waits for a percentile of the stopping probability, which limits how few iterations a token can run; a batched pass then runs every token for as long as the deepest one, which limits how little the pass can cost.

\begin{table*}[!t]

\begin{center}
\footnotesize
\setlength{\tabcolsep}{5pt}
\begin{tabular*}{\linewidth}{@{\extracolsep{\fill}}lccccl@{}}
\toprule
\multicolumn{1}{c}{\bf Tokens/pass} & \multicolumn{1}{c}{$R$}
& \multicolumn{1}{c}{$R\,\bar N$} & \multicolumn{1}{c}{\bf Sampled-depth there}
& \multicolumn{1}{c}{\bf Per-step $-$ sampled-depth} & \multicolumn{1}{c}{\bf Verdict} \\ \midrule
\multicolumn{6}{l}{\emph{370M, $\varepsilon{=}0.7$, $\bar N = 2.82$, BPB $1.0908$ --- same scale, model and threshold as $R$}} \\
$1$ (decode, $b{=}1$)          & 1.06 & 2.99 & 1.1082 & $\bf -0.0174$ & model ahead \\
$2048$ (prefill, $b{=}1$)      & 2.19 & 6.17 & 1.0889 & $+0.0019$ & inside floor \\
$65{,}536$ (prefill, $b{=}32$) & 2.32$^{\dagger}$ & 6.54 & 1.0886 & $+0.0022$ & inside floor \\[4pt]
\multicolumn{6}{l}{\emph{370M, $\varepsilon{=}0.95$, $\bar N = 1.62$, BPB $1.1086$ --- $R$ carried from $\varepsilon{=}0.7$}} \\
$1$ (decode, $b{=}1$)          & 1.06 & 1.72 & 1.1605 & $\bf -0.0519$ & model ahead \\
$2048$ (prefill, $b{=}1$)      & 2.19 & 3.55 & 1.1015 & $+0.0071$ & inside floor \\
$65{,}536$ (prefill, $b{=}32$) & 2.32$^{\dagger}$ & 3.76 & 1.0990 & $\bf +0.0096$ & baseline ahead \\[4pt]
\multicolumn{6}{l}{\emph{140M, $\varepsilon{=}0.7$, $\bar N = 3.11$, BPB $1.2196$ --- borrows $R$ across scale only}} \\
$1$ (decode, $b{=}1$)          & 1.06 & 3.30 & 1.2370 & $\bf -0.0174$ & model ahead \\
$2048$ (prefill, $b{=}1$)      & 2.19 & 6.81 & 1.2166 & $+0.0030$ & inside floor \\
$65{,}536$ (prefill, $b{=}32$) & 2.32$^{\dagger}$ & 7.22 & 1.2162 & $+0.0034$ & inside floor \\[4pt]
\multicolumn{6}{l}{\emph{770M, $\varepsilon{=}0.7$, \textbf{seed 42}: $\bar N = 2.74$, BPB $1.0174$ --- $R$ measured on this model at this scale}} \\
$1$ (decode, $b{=}1$)          & 1.10 & 3.01 & 1.0352 & $\bf -0.0178$ & model ahead \\
$2048$ (prefill, $b{=}1$)      & 2.35 & 6.44 & 1.0170 & $+0.0004$ & inside floor \\
$65{,}536$ (prefill, $b{=}32$) & 2.34 & 6.41 & 1.0170 & $+0.0004$ & inside floor \\
\bottomrule
\end{tabular*}
\end{center}
\caption{The $R$-priced comparison behind Figure~\ref{fig:priced}, at the timed widths; the $32$-sequence pass of Figure~\ref{fig:priced} is in Tables~\ref{tab:chain} and~\ref{tab:batcheddecode}. \emph{Sampled-depth there} is the sampled-depth
baseline read at $R\,\bar N$.
$R$: recovered from Table~\ref{tab:synchrony}'s per-step rows at 370M and $\varepsilon{=}0.7$ (Appendix~\ref{app:feasibility}); block headers
state what they borrow. \emph{Verdict} and bold: ``model ahead'' or ``baseline ahead'' if the margin exceeds twice the evaluation noise floor, ``inside floor'' otherwise. All margins are teacher-forced; with stopped tokens' keys and values hidden (simulated decode) the one-token rows change sign
(Appendix~\ref{app:decodematched}). $^{\dagger}$$R$ borrowed from a mixture-loss row.}
\label{tab:priced}
\end{table*}

\begin{figure}[t]
\centering
\includegraphics[width=\columnwidth]{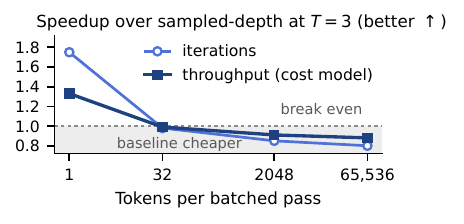}
\caption{\textbf{The speedup is largest when one token is generated at a time and is gone in large
batches.} Horizontal axis: how many tokens share one batched pass; $1$ and $32$ are decoding one or
$32$ sequences at once, and $2048$ and $65{,}536$ are processing one or $32$ prompts of $2048$ tokens.
Vertical axis: how many times faster the 370M per-step model at $\varepsilon{=}0.95$ is than the
sampled-depth baseline at $T{=}3$, whose teacher-forced quality it matches; above $1$ is faster. \emph{Iterations}: the speedup if only the core cost compute ($3$ divided by the iterations each pass runs). \emph{Throughput}: the speedup of the whole pass, as predicted by the cost model of Section~\ref{sec:feasibility}; it is smaller because the prelude and coda run once however many iterations the core runs (at one token per pass, $1.33\times$ against $1.75\times$). In a real decode with reused keys and values, the per-step model is
slightly worse than that baseline (Appendix~\ref{app:reuse}).}
\label{fig:deployable}
\end{figure}

\paragraph{Against a baseline a deployment can run.} The sampled-depth baseline at $T{=}3$ runs exactly three iterations,
since in a fixed-depth pass every token runs the same depth, and needs no interpolation. At $\varepsilon{=}0.95$ the
model is $0.0008$ BPB worse ($1.1086$ against $1.1078 \pm 0.0001$, three seeds each), inside the
floor. Counting the depth each pass runs and converting through the fitted $f$ of the cost model (Appendix~\ref{app:feasibility}), the model's speedup decays from
$1.33\times$ at one token per pass (teacher-forced quality) to $0.99\times$ at $32$ sequences, where the real decode counts $3.06$ iterations
a pass against the baseline's three, and falls further at prefill (Figure~\ref{fig:deployable}). In depth alone the one-token ratio would read $1.75\times$; the
prelude and coda sit outside the core.

\subsection{Simulated decode quality under exit}
\label{app:decodematched}

Hiding a stopped token's keys and values, as a one-token decoder must, reverses the per-step model's teacher-forced lead over the sampled-depth baseline (Table~\ref{tab:decodematched}). Except the real decodes below, every other quality figure under exit is teacher-forced, with an exited position still supplying keys
and values at the iterations it skips (Appendix~\ref{app:timing}). Here the evaluation hides them (simulated decode).
At iteration $r$, a position that exited before $r$ supplies no keys or values to any other position,
which is what a one-token decoder's cache holds. The prompt can be treated three ways. \emph{Keep-first}
(prompt at full depth): the first position of each sequence is never hidden, as when a decoder prefills a one-token prompt at full depth.
\emph{Drop} (prompt also exits): the exit rule applies to the first position too. \emph{512-token prompt}: the first $512$ positions
are fully computed and only later predictions are scored.

\paragraph{The first token under bfloat16 halting.} The halting probability is stored in bfloat16, a 16-bit number format, which rounds it.
A head that puts almost all its probability early then reaches a running total of exactly $1$ at iteration one: the first token of each sequence
exits at iteration one, and $\varepsilon = 0$ realizes $7.997$ steps rather than $8$; hiding that
token's deeper keys and values alone costs $0.010$ to $0.052$ BPB. Ouro's exit arithmetic, also run in bfloat16,
does the same (below). At true full depth in this evaluation, the model minus the baseline is $-0.0009$, $+0.0012$ and
$-0.0008$ BPB at 140M, 370M and 770M, inside the evaluation noise floor.

\paragraph{The per-step model's margin reverses in decode when stopped tokens' keys and values are hidden.} Table~\ref{tab:decodematched} gives the
result. Hiding barely moves the realized steps, so the comparison is at the same threshold,
and there the model that led by $0.020$--$0.025$ BPB teacher-forced trails the sampled-depth baseline by
$0.058$--$0.082$ at $\varepsilon = 0.7$ with a one-token prompt prefilled at full depth (keep-first). A decoder that
applies the exit rule to that prompt as well loses its deeper keys and values too, and the loss is $1.9$
to $3.4$ times larger (drop); that is what a real decode measures (below). Reusing the
stopped tokens' keys and values instead of hiding them keeps the lead (Appendix~\ref{app:reuse}).

\begin{table*}[!t]
\centering
\footnotesize
\setlength{\tabcolsep}{5pt}
\begin{tabular*}{\linewidth}{@{\extracolsep{\fill}}llccccccc@{}}
\toprule
& & \multicolumn{2}{c}{\bf Steps} & \multicolumn{4}{c}{\bf Model $-$ baseline at the model's own steps} \\
\bf Scale & $\varepsilon$ & teacher-forced & keep-first & teacher-forced & keep-first & drop & 512-token prompt \\ \midrule
140M & 0.7  & 3.113 & 3.112 & $-0.0202$ & $\mathbf{+0.0824}$\,\tiny{$\pm$.014} & $+0.1542$ & $\mathbf{+0.0433}$\,\tiny{$\pm$.0082} \\
     & 0.95 & 1.877 & 1.889 & $-0.0506$ & $\mathbf{+0.0764}$\,\tiny{$\pm$.015} & $+0.1233$ & $\mathbf{+0.0279}$\,\tiny{$\pm$.0087} \\[2pt]
370M & 0.7  & 2.827 & 2.833 & $-0.0221$ & $\mathbf{+0.0626}$\,\tiny{$\pm$.004} & $+0.1238$ & $\mathbf{+0.0348}$\,\tiny{$\pm$.0002} \\
     & 0.95 & 1.628 & 1.651 & $-0.0597$ & $\mathbf{+0.0439}$\,\tiny{$\pm$.004} & $+0.0643$ & $\mathbf{+0.0125}$\,\tiny{$\pm$.0006} \\[2pt]
770M & 0.7  & 2.752 & 2.767 & $-0.0246$ & $\mathbf{+0.0576}$\,\tiny{$\pm$.002} & $+0.1955$ & $\mathbf{+0.0276}$\,\tiny{$\pm$.0006} \\
     & 0.95 & 1.564 & 1.580 & $-0.0623$ & $\mathbf{+0.0400}$\,\tiny{$\pm$.004} & $+0.1336$ & $+0.0060$\,\tiny{$\pm$.0005} \\
\bottomrule
\end{tabular*}
\caption{The per-step $\beta{=}0.1$ model, CC, three seeds per cell (mean, and seed s.d.\ where
shown), against the sampled-depth baseline at the model's own realized steps. The last column keeps the first $512$ positions fully
computed and scores both only after them. \textbf{Bold} (keep-first and prompt columns): over twice the evaluation noise floor. The margin is positive in every
simulated decode cell and beyond twice the evaluation noise floor in all but one (770M, $\varepsilon = 0.95$, 512-token prompt). Margins are in BPB; a positive margin means the model is worse than the baseline.}
\label{tab:decodematched}
\end{table*}

\paragraph{Behind a 512-token prompt.} Decode is timed from a $512$-token prompt
(Appendix~\ref{app:timing}), so the last column of Table~\ref{tab:decodematched} keeps positions $0$
to $511$ visible as keys and scores only later predictions, for the model and the baseline alike, with
realized steps counted over the same positions. Teacher-forced margins on those positions match the full-sequence ones (within
$0.0006$ BPB), so the change comes from the prompt, not from which positions are scored. The longer prompt
narrows the reversal, to $0.028$--$0.043$ at $\varepsilon = 0.7$ and $0.006$--$0.028$ at
$\varepsilon = 0.95$; only at 770M and $\varepsilon = 0.95$ is it within twice the evaluation noise floor ($+0.006$, $1.3$ times the floor). It also narrows with scale at both
thresholds. The per-step model's speedup at one token per pass, the one setting where the cost of a batched pass does not apply
(Appendix~\ref{app:priced}), is therefore not available to these models when stopped tokens' keys and values are hidden. They were
trained with every iteration's keys and values present, which is consistent with a train--test mismatch rather than a
limit of stopping tokens early. Filling the missing entries instead of hiding them recovers it (Appendix~\ref{app:reuse}).

\paragraph{A real decode.} The simulated decode evaluation emulates decode inside a teacher-forced pass.
On the first $64$ validation sequences we also run the decode itself: the prompt in one pass under
the exit rule, then each later token alone through the model's cache, so a token that exits never
writes its deeper keys and values (Table~\ref{tab:cacheddecode}). Behind a $512$-token prompt the real
decode realizes the emulation's steps, with a margin $0.002$--$0.004$ BPB harsher (s.d.\ over seeds
$0.0007$--$0.0012$), so the reversal holds under a real cache. With a one-token prompt it equals
drop: the prompt token exits at iteration one under bfloat16, so its deeper keys and values never exist.

\begin{table}[!t]
\centering
\footnotesize
\setlength{\tabcolsep}{0.5pt}
\begin{tabular*}{\linewidth}{@{\extracolsep{\fill}}llccccc@{}}
\toprule
& & \multicolumn{2}{c}{\bf Steps} & \multicolumn{3}{c}{\bf Model $-$ baseline} \\
\bf Scale & \bf Prompt & real & emulated & real & emulated & drop \\ \midrule
140M & 512 & 3.006 & 3.006 & $+0.0701$ & $+0.0666$ & n/a \\
     & 1   & 2.979 & 3.005 & $+0.1941$ & $+0.1155$ & $+0.1940$ \\[2pt]
370M & 512 & 2.734 & 2.734 & $+0.0342$ & $+0.0319$ & n/a \\
     & 1   & 2.638 & 2.746 & $+0.1272$ & $+0.0581$ & $+0.1276$ \\[2pt]
770M & 512 & 2.651 & 2.650 & $+0.0223$ & $+0.0198$ & n/a \\
     & 1   & 2.805 & 2.685 & $+0.1834$ & $+0.0449$ & $+0.1841$ \\
\bottomrule
\end{tabular*}
\caption{Real decode against the emulation with keys and values hidden: per-step $\beta{=}0.1$ model,
$\varepsilon{=}0.7$, three seeds, first $64$ validation sequences ($98$k scored positions behind $512$
tokens, $131$k behind one). \emph{Emulated}: prompt keys and values kept at every depth. \emph{Drop}:
none kept. Margins against the sampled-depth baseline on the same positions; teacher-forced behind $512$ tokens,
$-0.018$, $-0.021$, $-0.026$. Full-split margins:
Table~\ref{tab:decodematched}. Margins are in BPB; a positive margin means the model is worse than the baseline.}
\label{tab:cacheddecode}
\end{table}

\paragraph{A batched real decode.} We also decode the full split $32$ sequences at a time in lockstep
through one cache, behind a $512$-token prompt, as a server running batched passes does: an exited token keeps
computing, and writing keys and values, until the deepest of the $32$ stops, and the depth of every
pass is recorded (Table~\ref{tab:batcheddecode}; the baselines scored on the same sequences). At $\varepsilon = 0.7$ a pass runs $4.37$--$4.71$ iterations, $R = 1.51$--$1.58$. At its own steps the model keeps most of its teacher-forced lead, since the pass supplies
the keys and values a one-token decode hides. Read at the depth the pass pays, the margin is
$-0.0004$, $-0.0010$ and $-0.0030$ BPB: a tie at 140M and 370M, and a win at 770M of eighteen times its
seed s.d.\ ($0.00017$; a $95\%$ $t$-interval over the three seeds is $0.0026$ to $0.0034$, and every seed
leads). Teacher-forced on the same sequences, every margin shifts $0.0022$--$0.0027$
toward the model, since there each position keeps its keys and values to the cap, not only to the depth its pass
ran. At $\varepsilon = 0.95$ the model loses at every scale, by $0.0019$ to $0.0048$.

\begin{table}[!t]
\centering
\footnotesize
\setlength{\tabcolsep}{1.5pt}
\begin{tabular*}{\linewidth}{@{\extracolsep{\fill}}lcccccc@{}}
\toprule
& & \bf Pass & & \multicolumn{3}{c}{\bf Model $-$ baseline at} \\ \cmidrule(l){5-7}
\bf Scale & $\bar N$ & \bf depth & $R$ & $\bar N$ & \bf pass depth & \bf TF \\ \midrule
\multicolumn{7}{@{}l}{$\varepsilon = 0.7$} \\
140M & 3.13 & 4.71 & 1.51 & $\mathbf{-0.0182}$ & $-0.0004$\,\tiny{$\pm$.0013} & $-0.0030$ \\
370M & 2.84 & 4.40 & 1.55 & $\mathbf{-0.0197}$ & $-0.0010$\,\tiny{$\pm$.0005} & $-0.0034$ \\
770M & 2.76 & 4.37 & 1.58 & $\mathbf{-0.0224}$ & $\mathbf{-0.0030}$\,\tiny{$\pm$.0002} & $-0.0053$ \\[2pt]
\multicolumn{7}{@{}l}{$\varepsilon = 0.95$} \\
140M & 1.89 & 3.28 & 1.74 & $\mathbf{-0.0473}$ & $\mathbf{+0.0048}$\,\tiny{$\pm$.0021} & $+0.0021$ \\
370M & 1.63 & 3.06 & 1.87 & $\mathbf{-0.0571}$ & $\mathbf{+0.0038}$\,\tiny{$\pm$.0006} & $+0.0013$ \\
770M & 1.57 & 3.03 & 1.93 & $\mathbf{-0.0594}$ & $\mathbf{+0.0019}$\,\tiny{$\pm$.0003} & $-0.0007$ \\
\bottomrule
\end{tabular*}
\caption{The per-step $\beta{=}0.1$ model under a batched real decode: the full split,
$32$ sequences per pass behind a $512$-token prompt, three seeds. \emph{Pass depth}: mean iterations
per pass, $R\,\bar N$ measured. Margins against the sampled-depth baseline on the same sequences;
\emph{TF}: teacher-forced at the pass depth. $\pm$: seed s.d.\ (at most $0.0033$ at
$\bar N$, $0.0017$ for TF); bold ($\bar N$ and pass-depth columns): over twice it.}
\label{tab:batcheddecode}
\end{table}

\paragraph{Ouro.} Ouro's released forward runs every loop step for every token
(Appendix~\ref{app:ouro}), so we run its loop again ourselves, one decoder layer at a time, hiding each exited token's deeper keys and
values at every later step, using its own exit arithmetic in bfloat16. We split each evaluation document into chunks of up to $2048$ tokens. On
Ouro's 2.6B model the first token of a chunk exits at step one in $1{,}244$ of $1{,}288$ chunks at
$\tau_{\text{Ouro}} = 0.05$, and keep-first recovers most of the quality that drop loses; on the 1.4B model it almost never
exits early and the two coincide. Two further ways of keeping an exited token visible, copy and reuse
(Appendix~\ref{app:reuse}), do not protect the first token, so on the 2.6B model their comparison with keep-first is confounded by it. On the 1.4B model,
copy (feeding the frozen exit state to every later layer) breaks
the model ($+1.56$ BPB at $\tau_{\text{Ouro}} = 0.05$), while reuse (keeping the keys and values from the exit step)
matches keep-first's quality ($+0.077$ against $+0.075$) at more realized steps ($2.83$ against
$2.65$).

\noindent At one token per pass, where the pass stops when its token exits and hiding the
deeper keys and values is the right model, the quality condition survives on both models under
keep-first (Table~\ref{tab:ouromasked}). In a batched pass, stopped tokens keep computing until the pass ends, so hiding their keys and values at every skipped depth would understate its quality. The batched figures of
Appendix~\ref{app:ouro} therefore stay teacher-forced. A quality-preserving threshold is not free at one
token per pass: at $\tau_{\text{Ouro}} = 0.9$ keep-first costs Ouro's 2.6B model $0.017$ BPB, outside the evaluation noise floor, where
teacher forcing reports none.

\begin{table}[!t]
\centering
\footnotesize
\setlength{\tabcolsep}{1.5pt}
\begin{tabular*}{\linewidth}{@{\extracolsep{\fill}}llccc@{}}
\toprule
\bf Model & \bf Variant & \bf Steps & \shortstack{\bf Change vs\\ \bf teacher-forced} & \bf Condition \\ \midrule
Ouro 2.6B & teacher-forced & 2.444 & ---       & $\mathbf{-0.300}$ \\
          & keep-first     & 2.475 & $+0.111$  & $\mathbf{-0.176}$ \\
          & drop           & 2.943 & $+0.264$  & $+0.181$ \\[2pt]
Ouro 1.4B & teacher-forced & 2.669 & ---       & $\mathbf{-0.278}$ \\
          & keep-first     & 2.654 & $+0.075$  & $\mathbf{-0.211}$ \\
\bottomrule
\end{tabular*}
\caption{Ouro at $\tau_{\text{Ouro}} = 0.05$ and one token per pass: the quality condition, the model minus the depth
curve at the realized steps (negative: the model passes). \emph{Keep-first}: the first token keeps its deeper keys and values; \emph{drop}: it hides them like any exited token. On the 1.4B model drop and keep-first
coincide.}
\label{tab:ouromasked}
\end{table}

\subsection{Filling the skipped keys and values}
\label{app:reuse}

A token that stops at iteration $r$ has no keys or values at the iterations after $r$. Hiding them
(Appendix~\ref{app:decodematched}) is one policy; a decoder can also fill them. We tried two, in this order.
\emph{Copy} computes the missing keys and values at every later layer from the token's state at the end of
iteration $r$, as early-exit decoders copy a state into skipped layers \citep{schuster2022confident}. No layer receives this frozen state as input in training: the first layer of each iteration reads the state after it is
combined with the input embedding, and later layers read the output of the layer before. \emph{Reuse}
keeps, at every later iteration and layer, the keys and values the token had at that layer in iteration
$r$. The weights are the same at every iteration, so these are entries the layer itself produced, and a key-value cache gets them by pointing the missing entries at the computed ones. This is the policy of \citet{geiping2025scaling}, following \citet{elbayad2019depth}. In a stack without shared weights, CALM (Confident Adaptive Language Modeling) found the opposite: copying keys and values into skipped layers hurts, while copying the hidden state does not \citep{schuster2022confident}; in our looped models, copying the hidden state (our \emph{copy} policy) is the one that fails. Both are scored like the
512-token-prompt column of Table~\ref{tab:decodematched}, after a $512$-token prompt.

Table~\ref{tab:reuse} gives the result. Copy is worse than hiding at 370M and 770M. Reuse keeps most of
the teacher-forced lead at every scale and threshold, and a real decode through the cache with reuse
gives the same margins as the emulation. On the same $64$ sequences the two differ by at most
$0.0003$ BPB; at 770M and $\varepsilon = 0.95$ that difference exceeds twice its seed s.d. The emulated columns of Table~\ref{tab:reuse} are on the full validation split and the
real column on the first $64$ sequences, so the two differ by as much as the subsets do. With reuse,
decoding $32$ sequences together gives each sequence the same BPB as decoding it alone.

\begin{table}[!t]
\centering
\footnotesize
\setlength{\tabcolsep}{0pt}
\begin{tabular*}{\linewidth}{@{\extracolsep{\fill}}llccccc@{}}
\toprule
& & & \multicolumn{3}{c}{\bf emulated, full split} & \bf real, $64$ seq. \\ \cmidrule(lr){4-6}\cmidrule(l){7-7}
\bf Scale & $\varepsilon$ & \bf TF & hidden & copy & reuse & reuse \\ \midrule
140M & 0.7  & $-0.021$ & $+0.043$ & $+0.015$ & $-0.019$ & $-0.016$ \\
     & 0.95 & $-0.050$ & $+0.028$ & $+0.012$ & $-0.039$ & $-0.044$ \\[2pt]
370M & 0.7  & $-0.022$ & $+0.035$ & $+0.098$ & $-0.020$ & $-0.019$ \\
     & 0.95 & $-0.059$ & $+0.013$ & $+0.209$ & $-0.050$ & $-0.050$ \\[2pt]
770M & 0.7  & $-0.025$ & $+0.028$ & $+0.237$ & $-0.022$ & $-0.025$ \\
     & 0.95 & $-0.062$ & $+0.006$ & $+0.375$ & $-0.052$ & $-0.054$ \\
\bottomrule
\end{tabular*}
\caption{Filling a stopped token's deeper keys and values: the per-step $\beta{=}0.1$ model minus the
sampled-depth baseline at the model's own realized steps, after a
$512$-token prompt, three seeds. \emph{TF}: teacher-forced. Seed s.d.\ at most $0.0087$ (hidden),
$0.098$ (copy), $0.0039$ (reuse). Margins are in BPB; a positive margin means the model is worse than the baseline.}
\label{tab:reuse}
\end{table}

\paragraph{Speed.} Reuse adds no measurable decode time. At one sequence per pass, after a
$512$-token prompt and timing the same $64$ generated tokens for every variant, reuse takes $1.005$ to
$1.021$ times as long as hiding at every scale and threshold. Measured against a different reference, the sampled-depth baseline run for
three iterations, at $\varepsilon = 0.7$ the model decodes at $0.88$--$0.89$, $0.98$--$1.01$ and
$1.02$--$1.05$ times the baseline's speed at 140M, 370M and 770M and is $0.016$, $0.013$ and $0.017$ BPB
better in a real decode; at $\varepsilon = 0.95$ it is $1.14$, $1.30$--$1.32$ and $1.35$--$1.36$ times
as fast and $0.007$, $0.006$ and $0.003$ BPB worse. Ranges are over three seeds; each comparison is a
median over $15$ interleaved rounds on one GPU; across the $15$ rounds a single comparison varied by $4$ to $41\%$ of its median, and we read ratios within about $5\%$ of $1$ as ties.

\paragraph{Gathering in decode.} With reuse, a batched decode can also gather: each iteration runs only the
sequences whose token is still running. Decoding $32$ sequences at $\varepsilon = 0.7$, this is $0.84$ to
$0.87$ times as fast as the batched pass at 370M and 770M, and $0.61$ to $0.64$ times as fast as the
baseline at three iterations; the batched pass matches the speed of the sampled-depth baseline run for five iterations ($1.01$--$1.03$
times), close to the $4.5$ to $4.8$ iterations it runs per pass on these timing tokens. An iteration over $32$ tokens costs about the same
however many of them run, so removing stopped ones saves little, and the gather adds work. This is one
implementation; a fused kernel could remove the overhead but not the iterations, so it could approach the
batched pass, not the baseline at three. The full-sequence result of Appendix~\ref{app:timing} is a
different setting (full sequences, not one token per pass) and does not carry over to decode.

\paragraph{Why reuse works here is open.} After a token stops, its keys and values still change. We measure the change at each later iteration as the size of the difference from the token's keys (or values) at its exit iteration, divided by the size of its keys (or values) at the later iteration. On average over the later iterations, the change is $13$--$18\%$ for keys and $20$--$26\%$ for values; for keys it rises from
$11$--$14\%$ one iteration later to $35$--$50\%$ six later. On Ouro at $\tau_{\text{Ouro}} = 0.05$ the change is about twice as
large ($31\%$ and $58$--$61\%$), and reuse does no better there than keeping the first token visible (above). But per-step models trained at
$\beta = 0.01$ change as much as Ouro ($32$--$53\%$ and $45$--$73\%$), with or without the attractor term (Section~\ref{sec:bg:arch}), and reuse still recovers $91$ to $102\%$ of the hidden
loss in them (140M and 370M, one seed each). The size of the change therefore does not explain when reuse
works.

\subsection{The mean is reachable, and that is the problem}
\label{app:sampled}
PonderNet's test-time rule~\citep[Sec.~2.5]{banino2021pondernet}, the \emph{sampled} rule,
draws each token's stopping iteration at random from $p_i$. On the mixture-loss model it realizes
$\mathbb{E}[N]$ and costs ten times the $0.02$ BPB by which a specialist trained at three iterations trails full depth (Section~\ref{sec:results:main}).

\paragraph{What the average costs to realize.} The mixture-loss model reaches its predicted average
under this rule and pays $0.198$--$0.234$ BPB for it (Table~\ref{tab:sampled}), measured against the
sampled-depth baseline run to the same depth, at two scales and three seeds. Reading out the exit state rather than the mixture costs a
further $0.05$--$0.06$ (Table~\ref{tab:sampled}): a sampled stop lands on a state the loss never scored, as in
Table~\ref{tab:decode}.

\begin{table*}[!tp]

\begin{center}
\small
\setlength{\tabcolsep}{5pt}
\begin{tabular*}{\linewidth}{@{\extracolsep{\fill}}llcccc@{}}
\toprule
\multicolumn{1}{c}{\bf Model} & \multicolumn{1}{c}{\bf Readout}
& \multicolumn{1}{c}{\bf Steps} & \multicolumn{1}{c}{\bf BPB $\downarrow$}
& \multicolumn{1}{c}{\bf Baseline} & \multicolumn{1}{c}{$\Delta$ $\downarrow$}
\\ \midrule
\multicolumn{6}{l}{\emph{Sampled rule: realized steps $= \mathbb{E}[N]$ by construction}} \\
140M mixture-loss              & mixture    & 3.830 & 1.4633 & 1.2291 & $+0.2342$ \\
140M mixture-loss              & exit state & 3.830 & 1.5262 & 1.2291 & $+0.2971$ \\
370M mixture-loss              & mixture    & 3.762 & 1.2966 & 1.0990 & $+0.1976$ \\
370M mixture-loss              & exit state & 3.762 & 1.3439 & 1.0990 & $+0.2449$ \\
140M per-step, $\beta{=}0.1$ & exit state & 4.335 & 1.2198 & 1.2244 & $-0.0046$ \\
770M per-step, $\beta{=}0.1$ & exit state & 3.993 & 1.0172 & 1.0242 & $-0.0070$ \\[3pt]
\multicolumn{6}{l}{\emph{Threshold rule at 140M, at comparable realized steps}} \\
140M mixture-loss ($\varepsilon{=}0.5$) & mixture    & 3.020 & 1.4493 & 1.2415 & $+0.2078$ \\
140M per-step ($\varepsilon{=}0.7$)   & exit state & 3.113 & 1.2196 & 1.2398 & $-0.0202$ \\
\bottomrule
\end{tabular*}
\end{center}
\caption{\textbf{What realizing the average costs.} PonderNet's sampled rule, whose realized steps
equal $\mathbb{E}[N]$ by construction, against the sampled-depth baseline at the same realized steps
(``Baseline''); the threshold rule at 140M is included for contrast. BPB; three seeds, three random draws of the stopping iteration per seed.
 Variation across seeds is $0.0070$--$0.0149$ on the mixture-loss models and $0.0003$--$0.0008$ on
the per-step models, against the mixture-loss effects of $0.198$ and above in the upper block.}
\label{tab:sampled}
\end{table*}

\paragraph{Neither rule rescues the mixture-loss model.} At its operating point the sampled rule is worse
than the threshold rule on both axes at once --- $3.83$ steps against $3.02$, and $+0.234$ against
$+0.208$. The choice for the mixture-loss model is between a rule that realizes the average and
costs $0.2$ BPB and a threshold rule that preserves the model only where it barely exits. For the per-step model
under a strong prior the sampled rule still beats the sampled-depth baseline at matched depth, by $0.0070$ at
770M and $0.0046$ at 140M, and it pays $1.2$ more iterations than the
threshold rule for the same quality ($1.2198$ against $1.2196$ at 140M). So the negative result
holds under both rules; teacher-forced, the per-step model with a strong prior beats the baseline under both.

\subsection{ACT's ponder weight selects a depth, not a policy}
\label{app:actponder}
ACT's depth varies across tokens at two of its seven ponder weights, and the sampled-depth baseline run shallower matches or overtakes it at
about $6.5$ iterations at both.

The ACT model reported throughout trains at $\tau_p = 10^{-4}$. The three weights of the selection in Appendix~\ref{app:training}
($10^{-4}$, $3\times10^{-3}$, $10^{-2}$, three seeds each) and four further 140M models, at
$3\times10^{-2}$, $10^{-1}$, $3\times10^{-1}$ and $1$, take the sweep to four orders of magnitude.

\paragraph{Two weights vary the depth, and the baseline run shallower matches both.} Table~\ref{tab:actponder} gives the
whole sweep. At $\tau_p = 10^{-2}$ the realized steps move from $7.35$ (per seed $7.03$ to $7.55$) to
$6.51$ as the threshold rises, and the model leads the depth curve by $0.011$--$0.012$ down to $6.98$ steps,
$0.009$ at $6.79$ and $0.002$ at $6.51$, inside the evaluation noise floor.
At $\tau_p = 3\times10^{-2}$ ACT also allocates per token: $\mathbb{E}[N]$ is $7.34$ with a per-token
s.d.\ of $1.32$, against $7.99$ at $10^{-4}$, and the
threshold rule moves the realized steps smoothly from $7.38$ to $1.01$. But its quality falls off with depth. Read against the same depth curve at the same realized steps, the model leads by
$0.009$--$0.010$ down to $6.79$ steps, trails by $0.026$ at $5.51$ and by $0.258$ at $3.01$: the crossover
is at about $6.5$ of eight iterations, so varying the depth buys about $1.5$ iterations of the eight
before the sampled-depth baseline run shallower overtakes it, worth at most about $1.2\times$, the architecture's
speedup limit $S_{\text{arch}}$.

\begin{table}[!t]

\begin{center}
\small
\setlength{\tabcolsep}{3pt}
\begin{tabular*}{\linewidth}{@{\extracolsep{\fill}}lcccc@{}}
\toprule
\multicolumn{1}{c}{$\tau_p$} & \multicolumn{1}{c}{$\varepsilon_{\text{ACT}}$}
& \multicolumn{1}{c}{\bf Steps} & \multicolumn{1}{c}{\bf BPB $\downarrow$}
& \multicolumn{1}{c}{\bf Baseline}
\\ \midrule
$10^{-4}$ (reported)$^{\ddagger}$ & $[0, 0.5]$ & 7.99 & 1.2027 & 1.2159 \\[3pt]
$3\times10^{-3}$$^{\ddagger}$ & $[0, 0.3]$ & 7.95--7.98 & 1.2032 & 1.2159 \\
$3\times10^{-3}$$^{\ddagger}$ & 0.5  & 7.862 & 1.2049 & 1.2160 \\[3pt]
$10^{-2}$$^{\ddagger}$ & 0    & 7.352 & 1.2048 & 1.2163 \\
$10^{-2}$$^{\ddagger}$ & 0.1  & 7.182 & 1.2049 & 1.2164 \\
$10^{-2}$$^{\ddagger}$ & 0.2  & 6.977 & 1.2056 & 1.2166 \\
$10^{-2}$$^{\ddagger}$ & 0.3  & 6.787 & 1.2079 & 1.2168 \\
$10^{-2}$$^{\ddagger}$ & 0.5  & 6.505 & 1.2154 & 1.2171 \\[3pt]
$3\times10^{-2}$ & 0    & 7.379 & 1.2061 & 1.2161 \\
$3\times10^{-2}$ & 0.05 & 7.135 & 1.2063 & 1.2163 \\
$3\times10^{-2}$ & 0.1  & 6.793 & 1.2079 & 1.2166 \\
$3\times10^{-2}$ & 0.2  & 5.505 & 1.2445 & 1.2187 \\
$3\times10^{-2}$ & 0.3  & 3.005 & 1.4992 & 1.2414 \\
$3\times10^{-2}$ & 0.5  & 1.005 & 1.8068 & 1.3813 \\[3pt]
$10^{-1}$ & 0            & 1.900 & 1.2531 & 1.2883 \\
$10^{-1}$ & $[0.05, 0.5]$ & 1.000 & 1.2531 & 1.3818 \\[3pt]
$3\times10^{-1}$ & 0            & 1.901 & 1.2546 & 1.2884 \\
$3\times10^{-1}$ & $[0.05, 0.5]$ & 1.000 & 1.2546 & 1.3818 \\[3pt]
$1$       & 0            & 1.357 & 1.2533 & 1.3447 \\
$1$       & $[0.05, 0.5]$ & 1.000 & 1.2533 & 1.3818 \\
\bottomrule
\end{tabular*}
\end{center}
\caption{\textbf{ACT's ponder weight selects a depth, not a policy.} Realized steps and BPB at 140M
on CC as the ponder weight $\tau_p$ and the eval-time threshold
$\varepsilon_{\text{ACT}}$ vary; ``Baseline'' is the sampled-depth baseline at the same
realized steps. $^{\ddagger}$Three seeds, Baseline from the three-seed curve;
other rows one seed. A row spanning an $\varepsilon_{\text{ACT}}$
interval moves by at most $0.03$ steps and $0.0001$ BPB within it.}
\label{tab:actponder}
\end{table}

At $10^{-1}$ and above, from $\varepsilon_{\text{ACT}} = 0.05$ on, the head halts every token at the first iteration, and the threshold is
inert: BPB is identical to four decimals across every threshold in $[0.05, 0.5]$. That constant
costs $0.050$--$0.052$ BPB against the reported model near the cap, about two and a half times the $0.02$ by which a specialist at three
iterations trails full depth (Section~\ref{sec:results:main}). So the weight passes from a constant depth near the cap ($10^{-4}$,
$3\times10^{-3}$), through two weights that vary it ($10^{-2}$,
$3\times10^{-2}$), to a constant depth of one.

\section{The loss, the prior, and their combination}

This section varies the prior weight $\beta$ under both losses and shows what a strong prior changes.

\subsection{Forced-depth curves under prior pressure}
\label{app:pressure}
Under the per-step loss, raising the prior weight reshapes the curve so that the rule can stop tokens early.

Table~\ref{tab:pressure} gives the forced-depth sweep at 140M as $\beta$ varies under the per-step loss, against the sampled-depth baseline at every depth, and \mbox{Table~\ref{tab:t1cells}} the $T{=}1$
cells under both losses. At $\beta = 0.01$ the per-step model is worse than mixture-loss at shallow
depth ($1.4544$ against $1.3722$ at $T{=}4$, seed $42$, not tabulated); at $\beta = 0.1$ it is ahead of the baseline at every
depth, resolved for $T \leq 3$.

\begin{table*}[!tp]

\begin{center}
\footnotesize
\setlength{\tabcolsep}{5pt}
\begin{tabular*}{\linewidth}{@{\extracolsep{\fill}}lccc@{}}
\toprule
\multicolumn{1}{c}{\bf Loss} & \multicolumn{1}{c}{$\beta{=}0.01$}
& \multicolumn{1}{c}{$\beta{=}0.1$} & \multicolumn{1}{c}{$\beta{=}1.0$}
\\ \midrule
\multicolumn{4}{l}{\emph{140M, BPB at forced $T=1$}} \\
Mixture & 1.8127\,\tiny{$\pm$.017} & 1.8508\,\tiny{$\pm$.044 $(+0.038)$} & 1.8648\,\tiny{$\pm$.044 $(+0.052)$} \\
Per-step   & 1.7111\,\tiny{$\pm$.003} & 1.3166\,\tiny{$\pm$.004 $\bf (-0.394)$} & 1.2608\,\tiny{$\pm$.000 $\bf (-0.450)$} \\[3pt]
\multicolumn{4}{l}{\emph{370M, BPB at forced $T=1$}} \\
Mixture & 1.6104\,\tiny{$\pm$.011} & 1.6083\,\tiny{$\pm$.027 $(-0.002)$} & 1.6415\,\tiny{$\pm$.026 $(+0.031)$} \\
Per-step   & 1.4302\,\tiny{$\pm$.011} & 1.1576\,\tiny{$\pm$.001 $\bf (-0.273)$} & 1.1225 \tiny{$\bf (-0.308)$} \\
$n$, mixture-loss / per-step & 3 / 3 & 3 / 3 & 3 / 1 \\
\bottomrule
\end{tabular*}
\end{center}
\caption{Forced-depth BPB at $T{=}1$ under both losses as the prior weight varies --- the cells
underlying Table~\ref{tab:twobytwo} (Section~\ref{sec:results:recipe}). Parenthesized values are the change from
$\beta = 0.01$. 140M is three seeds per cell. At 370M every printed change is the difference of the
two printed means, so the $\beta{=}1.0$ change compares one seed against a three-seed base; the
within-seed figure is in the text. \textbf{Bold}: beyond twice the seed s.d.}
\label{tab:t1cells}
\end{table*}

\begin{table*}[!tp]

\begin{center}
\footnotesize
\setlength{\tabcolsep}{4pt}
\begin{tabular*}{\linewidth}{@{\extracolsep{\fill}}lccccccc@{}}
\toprule
\multicolumn{1}{c}{\bf Model} & \multicolumn{1}{c}{$T{=}1$}
& \multicolumn{1}{c}{$T{=}2$} & \multicolumn{1}{c}{$T{=}3$}
& \multicolumn{1}{c}{$T{=}4$} & \multicolumn{1}{c}{$T{=}5$}
& \multicolumn{1}{c}{$T{=}6$} & \multicolumn{1}{c}{$T{=}8$}
\\ \midrule
Sampled-depth baseline      & 1.3818 & 1.2779 & 1.2415 & 1.2266 & 1.2201 & 1.2174 & 1.2155 \\
Per-step, $\beta{=}0.01$    & 1.7139 & 1.5612 & 1.5015 & 1.4544 & 1.4000 & 1.3327 & 1.2039 \\
Per-step, $\beta{=}0.1$     & 1.3169 & 1.2456 & 1.2250 & 1.2185 & 1.2164 & 1.2157 & 1.2154 \\
Per-step, $\beta{=}1.0$     & 1.2608 & 1.2344 & 1.2278 & 1.2261 & 1.2256 & 1.2255 & 1.2255 \\[4pt]
$\Delta$, $\beta{=}0.01$    & $+0.3321$ & $+0.2833$ & $+0.2600$ & $+0.2278$ & $+0.1799$ & $+0.1153$ & $\bf -0.0116$ \\
$\Delta$, $\beta{=}0.1$     & $\bf -0.0649$ & $\bf -0.0323$ & $\bf -0.0165$ & $-0.0081$ & $-0.0037$ & $-0.0017$ & $-0.0001$ \\
$\Delta$, $\beta{=}1.0$     & $\bf -0.1210$ & $\bf -0.0435$ & $\bf -0.0137$ & $-0.0005$ & $+0.0055$ & $+0.0081$ & $+0.0100$ \\
\bottomrule
\end{tabular*}
\end{center}
\caption{Forced-depth BPB at 140M on CC as the prior weight $\beta$ varies under the per-step loss,
and the difference against the sampled-depth baseline at each depth. Weights are unchanged between columns; only the forced depth $T$ changes. Seed $42$ for every
row except $\beta{=}1.0$ (three-seed means) and the baseline (three-seed mean). \textbf{Bold}: the model ahead by more than twice the evaluation noise floor, the criterion since most rows are single-seed. The resolved band at $\beta{=}0.1$ is $T \leq 3$.}
\label{tab:pressure}
\end{table*}

\paragraph{The drift.} Under $\beta = 0.01$ the per-step model's training-time $\mathbb{E}[N]$ climbs
from about $3.0$ to $4.87$, $5.09$ and $5.55$ at $1.5$k, $4$k and $9$k steps, past the prior's mean of
$3.11$; at 370M the mixture-loss model at the same weight settles at $3.47$ and the per-step model at
$5.07$. The 140M trajectory is seed $42$. The per-step model's exit rule at $\beta = 0.01$ still runs tokens almost to the cap: at 370M, even at the largest threshold, $\varepsilon = 0.95$, the per-step model at $\beta = 0.01$ realizes $6.23$ to $6.50$ of $8$ steps over its three seeds (exit-state readout).

\paragraph{The 370M interaction.} With three seeds per mixture-loss cell the $T{=}1$ response is
$-0.002$ and $+0.031$ against a standard error of $0.016$: mixture-loss at 370M is inert to the prior at
one iteration. Differenced within a seed, the per-step model moves
$-0.273 \pm 0.010$ at $\beta{=}0.1$ and $-0.303$ at $\beta{=}1.0$ (seed $42$). Raising $\beta$ also
tightens the per-step seed s.d.\ eightfold, $0.0114$ to $0.0014$, where mixture-loss spans
$0.011$--$0.027$. This is what one would expect if a strong prior trains the early iterations under the per-step loss only, but we did not test this.
At other depths the prior does move mixture-loss: it helps by $0.019$--$0.032$ at $T = 4$--$6$ and
hurts by $0.027$--$0.036$ at $T{=}2$. Table~\ref{tab:curves370} gives the 370M curves at $T = 1$, $3$, $5$ and $8$.

\begin{table}[!t]
\centering
\setlength{\tabcolsep}{2pt}
\small
\begin{tabular*}{\linewidth}{@{\extracolsep{\fill}}lcccc@{}}
\toprule
\multicolumn{1}{c}{\bf Model} & \multicolumn{1}{c}{$T{=}1$} & \multicolumn{1}{c}{$T{=}3$}
& \multicolumn{1}{c}{$T{=}5$} & \multicolumn{1}{c}{$T{=}8$} \\ \midrule
Sampled-depth baseline      & 1.2214 & 1.1079 & 1.0912 & 1.0875 \\
Per-step, $\beta{=}0.01$    & 1.4302 & 1.2505 & 1.1817 & 1.0924 \\
Per-step, $\beta{=}0.1$     & 1.1576 & 1.0935 & 1.0880 & 1.0874 \\
Per-step, $\beta{=}1.0$     & 1.1225 & 1.0955 & 1.0940 & 1.0939 \\
Mixture, $\beta{=}0.01$  & 1.6104 & 1.2765 & 1.1844 & 1.0807 \\
Mixture, $\beta{=}0.1$   & 1.6083 & 1.2764 & 1.1588 & 1.0813 \\
Mixture, $\beta{=}1.0$   & 1.6415 & 1.2749 & 1.1527 & 1.0806 \\
\bottomrule
\end{tabular*}
\caption{Forced-depth BPB at 370M under both losses and three prior weights. Model rows are three seeds, except per-step $\beta{=}1.0$ (seed $42$); the baseline is seed $42$.}
\label{tab:curves370}
\end{table}

\paragraph{Differences, not levels.} Each $\Delta$ in Table~\ref{tab:pressure} is taken within one
evaluation, whose three baseline seeds agree to $0.0006$ past one iteration. The baseline's BPB at the cap is $1.2155$ in that table but $1.2193 \pm 0.0026$ in Table~\ref{tab:main},
because the two were evaluated in different sessions; so we subtract only within one table.

\subsection{The loss--prior interaction on FW}
\label{app:fwinteraction}
The interaction replicates on a second corpus at two to four iterations, the range where an exit rule works, though not at one iteration.

Table~\ref{tab:fwinteraction} repeats the $2\times2$ at 140M on FW, seed $42$, where the evaluation noise floor is $0.0045$. The
interaction, $\beta{=}0.01 \to 0.1$ per-step minus mixture-loss, is $-0.344$, $-0.338$ and $-0.310$ at
$T = 2$, $3$, $4$ on FW against $-0.315$, $-0.256$ and $-0.208$ on CC (three seeds, not tabulated). On FW at $T = 2$--$4$ the prior moves
the per-step model by $0.32$--$0.34$ BPB and mixture-loss by at most $0.008$. At $T = 2$ and $4$ only
per-step at $\beta = 0.1$ beats the sampled-depth baseline, and the penalty at $\beta = 0.01$ replicates: at
$\beta = 0.01$ per-step is worse than mixture-loss at $T = 2$--$4$.
What does not replicate is $T{=}1$: on FW both losses improve by $0.37$--$0.40$, so the
asymmetry CC shows there is absent.

\begin{table*}[!tp]

\begin{center}
\footnotesize
\setlength{\tabcolsep}{3pt}
\begin{tabular*}{\linewidth}{@{\extracolsep{\fill}}lccccc@{}}
\toprule
\multicolumn{1}{c}{\bf Model} & \multicolumn{1}{c}{$T{=}1$}
& \multicolumn{1}{c}{$T{=}2$} & \multicolumn{1}{c}{$T{=}3$}
& \multicolumn{1}{c}{$T{=}4$} & \multicolumn{1}{c}{$T{=}8$}
\\ \midrule
Sampled-depth & -- & 0.9439 & -- & 0.9122 & 0.9054 \\[3pt]
Mixture $0.01$ & 1.7119 & 1.1462 & 1.0577 & 1.0155 & 0.8997 \\
Mixture $0.1$ & 1.3168 \tiny{$(-.395)$} & 1.1487 \tiny{$(+.003)$}
  & 1.0601 \tiny{$(+.002)$} & 1.0072 \tiny{$(-.008)$} & 0.8995 \tiny{$(-.000)$} \\[3pt]
Per-step $0.01$ & 1.3446 & 1.2675 & 1.2464 & 1.2241 & 0.8969 \\
Per-step $0.1$ & 0.9766 \tiny{$(-.368)$} & 0.9263 \tiny{$(-.341)$}
  & 0.9108 \tiny{$(-.336)$} & 0.9057 \tiny{$(-.318)$} & 0.9035 \tiny{$(+.007)$} \\
\bottomrule
\end{tabular*}
\end{center}
\caption{Loss $\times$ prior weight at 140M on FW, forced-depth BPB, seed $42$. Row labels give $\beta$. Parenthesized
values are the change from $\beta = 0.01$ at that depth. At $T=8$ the changes are at most $0.007$.}
\label{tab:fwinteraction}
\end{table*}

\subsection{The specialist at the cap, and allocation}
\label{app:fixedcontrol}
At full depth, the per-step model's margin over the sampled-depth baseline comes from how that baseline was trained: randomized depth and four-iteration gradient reach.

Table~\ref{tab:fixedcontrol} gives the per-seed comparison of the per-step model with a specialist trained at $T = 8$. The sampled-depth baseline randomizes its depth and backpropagates through four iterations where the halting model reaches all
eight. A specialist trained at $T = 8$ carries neither: gradients through every iteration, no
halting head and no halting loss. At $+0.0001 \pm 0.0017$, with the sign flipping across seeds,
the model and the specialist are indistinguishable. ACT trains as the halting models do, at fixed $T = 8$ with gradients through all eight, so the
halting-versus-ACT contrast is unaffected.

\paragraph{Gradient reach at 770M.} We retrain the 770M sampled-depth baseline with gradients through
the last eight iterations instead of four ($\mu_{\text{bwd}}{=}8$, three seeds) and measure it in one
session with the published baseline. On the full split its BPB is lower at every depth: by
$0.0005$--$0.0006$ at two and three iterations, where the exit rule lands (about one seed s.d.), and by
$0.0017$ at the cap; its depth curve's span from
$t{=}3$ to $8$ is $0.0203 \pm 0.0003$ against $0.0193 \pm 0.0002$. In the batched real decode of
Section~\ref{sec:batching:pass}, rerun in one session with both baselines, the per-step model at
$\varepsilon = 0.7$ leads it by $0.0022 \pm 0.0001$ BPB at $R\,\bar N$, against $0.0030 \pm 0.0001$ (seed s.d.) over
the published baseline, which reproduces Table~\ref{tab:chain}; read at $\bar N$ in the same decode, the
leads are $0.0221$ and $0.0223$.
The lead at $R\,\bar N$ holds against each of the retrained baseline's three seeds ($0.0017$ to $0.0029$).

\begin{table}[!t]

\begin{center}
\small
\setlength{\tabcolsep}{3pt}
\begin{tabular*}{\linewidth}{@{\extracolsep{\fill}}lccc@{}}
\toprule
\multicolumn{1}{c}{\bf Seed} & \multicolumn{1}{c}{\bf Per-step}
& \multicolumn{1}{c}{\bf Specialist $T{=}8$} & \multicolumn{1}{c}{$\Delta$}
\\ \midrule
42  & 1.2039 & 1.2047 & $-0.0008$ \\
7   & 1.2056 & 1.2035 & $+0.0021$ \\
123 & 1.2055 & 1.2065 & $-0.0010$ \\[2pt]
Mean & 1.2050\,\tiny{$\pm$.0010} & 1.2049\,\tiny{$\pm$.0015} & $+0.0001$\,\tiny{$\pm$.0017} \\
\bottomrule
\end{tabular*}
\end{center}
\caption{The per-step model ($\beta{=}0.01$) against a specialist trained at
$T=8$, 140M, three seeds each, same data and schedule. $\Delta$ is the model minus the specialist (negative favors the model); mean $\pm$ s.d.\ over seeds.}
\label{tab:fixedcontrol}
\end{table}

\paragraph{Two references for allocation.} Against a uniform depth of equal compute the exit rule adds
$21$ to $30\%$ of the lead (Appendix~\ref{app:adaptexit}). A uniform depth does not test whether the head tracks
difficulty, since a mix of depths can beat it whenever per-token loss curves differ in shape;
shuffling the depths does.

\paragraph{Shuffled depths.} Keep the depths the exit rule chose, shuffle which token gets which,
and re-score. Realized steps are preserved exactly and each token
keeps its byte count, so any difference is the assignment. On this model at $\varepsilon = 0.7$
($n=3$, 140M, $\beta = 0.1$) the assignment is worth $0.0083 \pm 0.0004$ BPB, positive on all three
seeds ($0.0079$--$0.0086$), varying by under $0.0001$ over twenty shuffles (Table~\ref{tab:permute}). Section~\ref{sec:results:lead} reads it as a share of the lead. Ouro's gate is worth
$0.30$ BPB over the same mean depth assigned without regard to the token (Appendix~\ref{app:ouro}).

\paragraph{Which tokens get depth.} We join each token's loss at every forced depth to the step the
exit rule chose. The rule gives the most depth to the highest-loss quintile (5), though depth helps
the middle one (3) most (Table~\ref{tab:alloctoken}). Rare tokens get more depth
than frequent ones ($3.34$
against $3.08$ at 140M, $2.97$ against $2.68$ at 770M). At 140M, tokens the exit rule stops at three
iterations improve by $0.028$ nats per token on the third and by $0.011$ over the five after it (three-seed
mean). The chosen step correlates with loss at the cap (Spearman $0.48$--$0.56$) more than with
the improvement from depth ($0.21$--$0.23$).

\begin{table}[t]
\centering
\footnotesize
\setlength{\tabcolsep}{2pt}
\begin{tabular*}{\linewidth}{@{\extracolsep{\fill}}llccccc@{}}
\toprule
& & \multicolumn{5}{c}{\bf Quintile of loss at the cap} \\
\bf Scale & & 1 & 2 & 3 & 4 & 5 \\ \midrule
140M & step & 2.28 & 3.08 & 3.33 & 3.40 & 3.48 \\
     & improvement & 0.06 & 0.29 & 0.34 & 0.24 & 0.12 \\[2pt]
370M & step & 2.06 & 2.69 & 3.06 & 3.11 & 3.21 \\
     & improvement & 0.01 & 0.15 & 0.28 & 0.20 & 0.09 \\[2pt]
770M & step & 2.03 & 2.54 & 2.98 & 3.06 & 3.16 \\
     & improvement & 0.01 & 0.12 & 0.26 & 0.21 & 0.09 \\
\bottomrule
\end{tabular*}
\caption{Where the per-step model ($\beta{=}0.1$, $\varepsilon{=}0.7$) spends depth (quintile 1:
easiest). \emph{Step}: realized steps. \emph{Improvement}: the loss drop from one iteration to the cap, in nats per token. CC; three-seed means.}
\label{tab:alloctoken}
\end{table}

\subsection{Why the mixture scores no single iteration}
\label{app:objectives}
Unless $p_i$ concentrates on one iteration, the mixture loss scores no single iteration; this follows from its form.

\paragraph{The one exception.} Where $p_i$ concentrates on a single $t$ the mixture \emph{is}
$h_{i,t}$ (Section~\ref{sec:bg:props}). The mixture-loss models are far from it: $\mathbb{E}[N]$ is $3.6$--$3.8$ while
a threshold needing $90\%$ of the probability stops tokens only at $7.6$--$7.9$ steps.

\paragraph{The mixture gap is not a usable measure.} The gap
$\sum_t p_{i,t}\mathrm{CE}(h_{i,t}) - \mathrm{CE}(\bar h_i)$ would be a signed measure of how far
the iterations disagree only under a linear readout, and ours is an $n_c$-layer coda. Measured on the per-step models, it
changes sign during training: $+0.0046$ nats at $42\%$ of training at 140M, and negative at
the end ($-0.0023$ at 140M, $-0.0220$ at 370M), which a linear readout rules out: cross-entropy is convex, so the mixture's loss could not exceed the average of the iterations' losses. So
Appendix~\ref{app:decode} measures the readout penalty directly instead.

\subsection{The readout penalty in full}
\label{app:decode}
Reading one iteration instead of the mixture costs $0.17$--$0.20$ BPB, and the per-step loss removes it.

Table~\ref{tab:decode} is measured at $\varepsilon = 0$, so the state read is $h_{i,T}$, the
last iteration, which carries whatever probability is left (Section~\ref{sec:bg:arch}). Every mixture-loss row is the $\beta = 0.01$ model at its scale; the
per-step rows are $\beta = 0.01$ at 140M and 370M and $\beta = 0.1$ at 770M, the only per-step model
there, so the 770M removal changes the prior as well and the two smaller scales carry the claim.
Every row is three seeds.

\paragraph{The state-at-cap readout is poorly determined, and the mixture is not.} Across seeds the
mixture moves $0.0006$, $0.0001$, $0.0012$ and $0.0017$ BPB at the four scales and the state at the cap
$0.040$, $0.015$, $0.006$ and $0.016$. Nothing scores a single iteration under mixture loss, so what is read off one is whatever the seed left. The effect is five to thirty times
that noise at every scale.

\paragraph{The prior grows the mixture-loss penalty and leaves the per-step one near zero.} At 140M (seed $42$, one session), as $\beta$ goes
$0.01 \to 0.1 \to 1.0$ the mixture-loss penalty grows $+0.164 \to +0.553 \to +0.851$ BPB while the
per-step penalty stays within $0.0037$ of zero. Tightening the prior makes a single iteration less
readable for the mixture-loss model and leaves it readable for the per-step model.

\section{Concurrent work}
\label{app:related}
\citet{liu2026decodingloopedtransformersbetter} reduce compute without a halting head: every token runs the same, smaller number of iterations, and contrastive decoding recovers the quality that the shorter run loses. Because every token stops at the same iteration, the cost of Section~\ref{sec:batching:pass} does not arise.

\citet{schwethelm2026isodepth} point out that tokens stopping at different iterations break batched decoding. \citet{schwethelm2026depthadaptive} remove this cost for Ouro \citep{zhu2025scaling} and the model of \citet{geiping2025scaling} by filling the places of stopped tokens with tokens from other sequences; this does not change the iteration at which a token stops, so the saving is still bounded by the realized steps. They also report that Ouro's 1.4B model keeps its accuracy on grade-school math problems (GSM8K) down to about $2.5$ of its $4$ iterations; our criterion is BPB instead, and at the threshold where Ouro's 2.6B model keeps its BPB, it runs $3.55$ of its $4$ iterations (Appendix~\ref{app:ouro}).

\citet{mishra2026dynamic} reuse a stopped token's keys and values at the deeper iterations, as in Section~\ref{sec:batching:onetoken}. In agreement with Section~\ref{sec:batching:pass}, at a batch of $32$ sequences their method is slower than running every token for a fixed four iterations. Their criterion is task accuracy rather than BPB.

\citet{popescu2026adaptive} independently describe the mechanism behind our per-step result: when one exit distribution both decides when a token stops and weights the per-iteration losses in training, choosing the exit and shaping the iterations become entangled.

\citet{logan2026fixedpoint} stops each token once its output stops changing, without training a head. A batched pass still waits for its deepest token, as \citet{logan2026fixedpoint} also notes.

\citet{monea2026shared} let later iterations read the keys and values of the first iteration, which saves memory. Every token runs the same number of iterations, so the cost of tokens stopping at different iterations does not arise.

\end{document}